%% file: main.tex
\documentclass{article}
\usepackage{iclr2027_conference,times}

\input{math_commands.tex}

\usepackage{hyperref}
\usepackage{url}
\usepackage{graphicx}
\usepackage{caption}
\usepackage{subcaption}  %
\usepackage{enumitem}
\usepackage{amsmath}
\usepackage{multirow}
\usepackage{booktabs}
\usepackage{pifont}
\usepackage{fvextra}
\usepackage{float}  %
\usepackage{wrapfig}  %

\usepackage{xcolor}
\definecolor{deepred}{HTML}{940000}
\definecolor{citepurple}{rgb}{0.4,0.15,0.95}
\hypersetup{
    colorlinks=true,
    linkcolor=deepred,
    citecolor=citepurple,
    urlcolor=citepurple
}
\usepackage[capitalize,noabbrev]{cleveref}

\usepackage[toc,page,header]{appendix}
\usepackage[nohints]{minitoc}

\renewcommand \thepart{}
\renewcommand \partname{}

\title{RoboJudge: Multimodal Language Models as\\Judges for Embodied Video Generation}

\author{\fontsize{9.5pt}{\baselineskip}\selectfont
Siyuan Ma\textsuperscript{1,*,\textdagger}~~~~Yu Qi\textsuperscript{3,*}~~~~Cheng Ye\textsuperscript{4,*}~~~~Ruotian Peng\textsuperscript{2,5}~~~~Xinyi Xu\textsuperscript{3}~~~~Zitiantao Lin\textsuperscript{3}\\
\fontsize{9.5pt}{\baselineskip}\selectfont \textbf{Zedong Cai\textsuperscript{6}~~~Yijian Huang\textsuperscript{6}~~~Weiyang Liu\textsuperscript{1}~~~Siyuan Wang\textsuperscript{1,\textdaggerdbl}~~~Yandong Wen\textsuperscript{2,\textdaggerdbl}}\\[1.25mm]
\fontsize{9.5pt}{\baselineskip}\selectfont
\textsuperscript{1}The Chinese University of Hong Kong \quad
\textsuperscript{2}Westlake University \quad
\textsuperscript{3}Northeastern University\\[0.25mm]
\fontsize{9.5pt}{\baselineskip}\selectfont \textsuperscript{4}University of California, San Diego \quad
\textsuperscript{5}Zhejiang University \quad
\textsuperscript{6}Independent\\[0.25mm]
\fontsize{9.5pt}{\baselineskip}\selectfont \textsuperscript{*}Equal contribution \quad
\textsuperscript{\textdagger}Project lead \quad
\textsuperscript{\textdaggerdbl}Corresponding author
}

\newcommand{\cmark}{\ding{51}} %
\newcommand{\xmark}{\ding{55}} %

\iclrfinalcopy

\begin{document}

\maketitle
\doparttoc
\faketableofcontents

\input{Sections/abstract}

\begin{figure}[t]
\centering
\vspace{-1mm}
\includegraphics[width=1\linewidth]{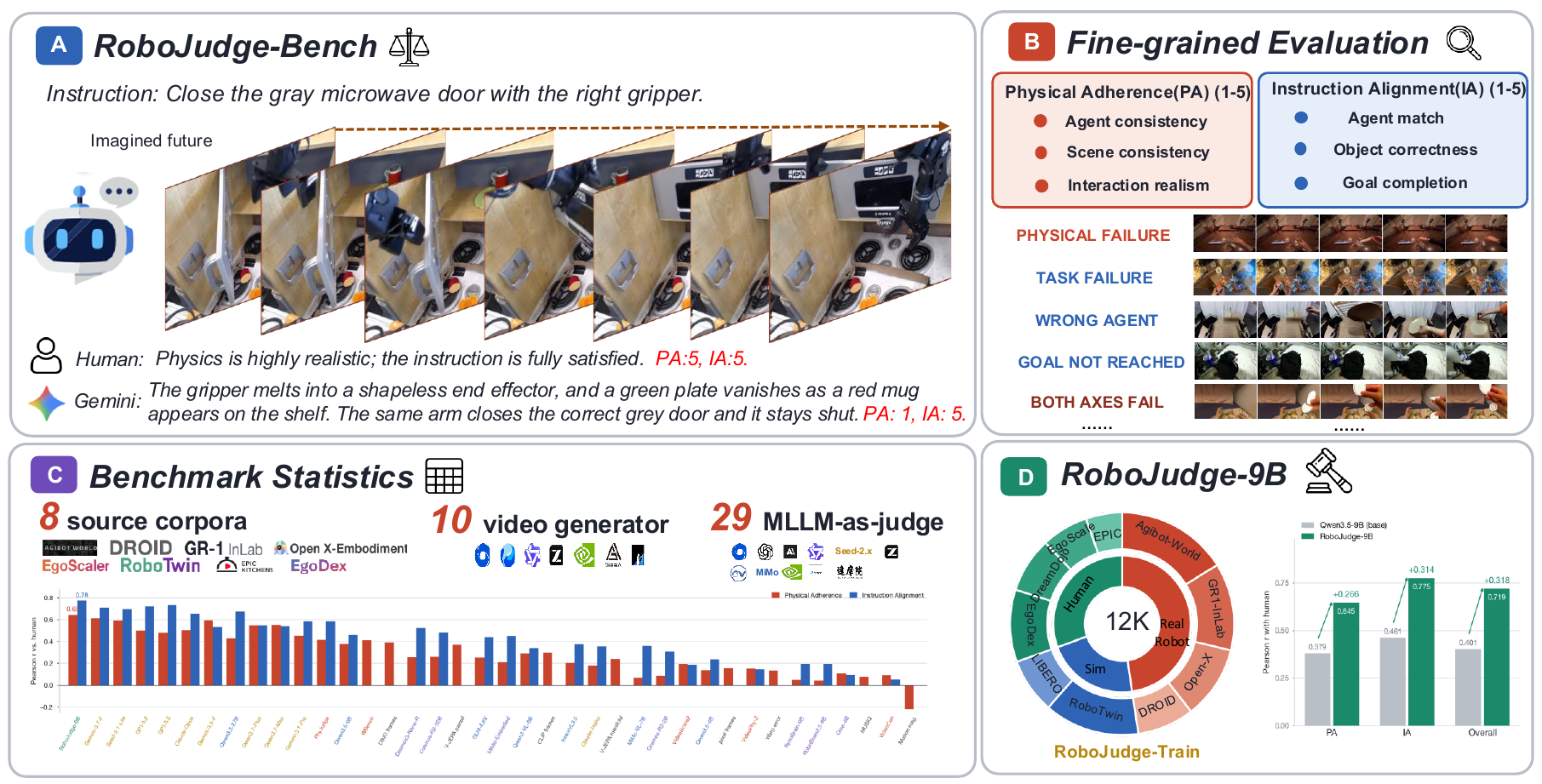}
\vspace{-5mm}
\caption{\textbf{Overview of \textsc{RoboJudge}.}
(A) We propose \textsc{RoboJudge-Bench}, which evaluates MLLMs as judges of embodied videos.
(B) \textit{Physical Adherence} and \textit{Instruction Alignment} serve as the two core evaluation dimensions for embodied judge models.
(C) We benchmark 29 existing model judges,
(D) and construct \textsc{RoboJudge-Train}, a 12K-video dataset for judge training, and fine-tune \textsc{RoboJudge-9B}, which achieves the strongest overall agreement among model judges.}
\label{fig:overview}
\vspace{-2mm}
\end{figure}

\section{Introduction}
\input{Sections/introduction}

\section{Related Work}
\input{Sections/related_work}

\section{RoboJudge-Bench}
\label{sec:judgebench}
\input{Sections/robojudge_bench}

\section{RoboJudge-9B}
\input{Sections/robojudge_9b}

\section{Experiments and Results}
\label{sec:experiment}
\input{Sections/experiments}

\section{Conclusion}
\input{Sections/conclusion}

\bibliography{references}
\bibliographystyle{iclr2027_conference}
\newpage
\appendix
\onecolumn
\addcontentsline{toc}{section}{Appendix}
\renewcommand \thepart{}
\renewcommand \partname{}
\part{\Large{\centerline{Appendix}}}
\parttoc

\newpage

\raggedbottom  %
\input{appendix/annotation}
\input{appendix/evaluation_protocol}

\input{appendix/data_statistics}

\input{appendix/video_generation_prompt_engineering}
\input{appendix/additional_analysis}
\input{appendix/additional_examples}
\input{appendix/experiment_details}

\end{document}

%% file: math_commands.tex
\usepackage{amsmath,amsfonts,bm}

\def\eqref#1{equation~\ref{#1}}

\def\1{\bm{1}}

\DeclareMathAlphabet{\mathsfit}{\encodingdefault}{\sfdefault}{m}{sl}
\SetMathAlphabet{\mathsfit}{bold}{\encodingdefault}{\sfdefault}{bx}{n}

%% file: Sections/abstract.tex
\begin{abstract}

Video generation models are promising foundations for embodied world models,
but only if their generated rollouts are physically valid and faithful to task
instructions. Multimodal large language models (MLLMs) are increasingly used
to judge these properties, yet their agreement with human judgments on
embodied videos has not been systematically evaluated. To address this gap,
we introduce \textbf{\textsc{RoboJudge-Bench}}, a benchmark for evaluating
MLLMs as judges of embodied video generation. The benchmark measures two
fundamental dimensions, \textit{Physical Adherence} and \textit{Instruction
Alignment}, across 800 human-annotated videos from 8 source corpora and 10
video generators, spanning grippers, human hands, and dexterous hands.
Experiments reveal uneven alignment between existing judges and human ratings:
general-purpose MLLMs show lower agreement on physical adherence, while
specialized video judges designed for general-domain physical reasoning do not
consistently transfer to embodied scenarios. Motivated by these findings, we
construct \textbf{\textsc{RoboJudge-Train}}, comprising 12K videos annotated with 25K human-verified scores
and model-assisted rationales for Physical Adherence and Instruction Alignment. We fine-tune
a 9B MLLM on this dataset with supervised learning followed by reinforcement
learning, yielding \textbf{\textsc{RoboJudge-9B}}. Among the model judges
evaluated on \textsc{RoboJudge-Bench}, \textsc{RoboJudge-9B} achieves the highest
overall agreement, reaching a Pearson correlation of $0.719$ with a statistically
significant 0.089 gain over Gemini-3.7-Flash, the best-performing proprietary
baseline in our evaluation. Training on \textsc{RoboJudge-Train} also yields gains on external
benchmarks, including VideoPhy2 and Physion-Eval.
We further use \textsc{RoboJudge-9B} as a downstream evaluator to benchmark
video generation models according to Physical Adherence and Instruction Alignment.
\begin{table}[H]
    \centering
    \begin{tabular}{l@{\textbf{:} }l}
        \textbf{Project page}        & \url{https://spherelab.ai/RoboJudge}
    \end{tabular}
\end{table}
\end{abstract}

%% file: Sections/introduction.tex
Recent advances in video generation have positioned generative models as a
promising foundation for embodied world models~\citep{worldmodelbench,ewmbench,
shang2026worldarena,jiang2026robowm,li2026causal}. Unlike general
video generators, embodied world models must predict how the physical world
evolves under robot actions or task instructions, providing imagined rollouts
for planning and policy learning. These rollouts therefore need to be not only
visually plausible, but also physically valid and instruction-consistent,
creating a pressing need for reliable evaluation.

Evaluating embodied generated videos is particularly challenging because they
require fine-grained assessment of physical interactions and task-level
correctness, while open-ended task execution makes reference-based similarity
intrinsically unreliable. Pixel, perceptual, and representation metrics capture
visual or semantic similarity to reference trajectories~\citep{huynh2008scope,
bakurov2022structural,zhang2018unreasonable,radford2021clip,caron2021dino,
bardes2024vjepa}, but cannot disentangle physical and task-level failures.
MLLM-based judges offer stronger multimodal reasoning~\citep{zheng2023judging,
mllmasajudge}, yet general-purpose MLLMs lack embodied-specific calibration~\citep{qi2026bear}.
Meanwhile, specialized video judges target general-domain quality or physical
commonsense~\citep{videoscore,videoscore2,bansal2025videophy,
lin2026phyground}, rather than fine-grained manipulation dynamics, contact
relationships, and task completion.

To address this gap, we introduce \textbf{\textsc{RoboJudge-Bench}}, a
benchmark for evaluating MLLMs as judges of embodied world model rollouts
along two critical axes: \textit{Physical Adherence (PA)} and \textit{Instruction
Alignment (IA)}. Specifically, PA covers \textit{agent consistency},
\textit{scene consistency}, and \textit{interaction realism}; IA covers
\textit{agent match}, \textit{object correctness}, and \textit{goal
completion}. The benchmark comprises 800 rollouts with instructions from
8 embodied source corpora and videos generated by 10 video generation models,
spanning parallel-jaw grippers, dexterous hands, and human hands across indoor
and outdoor activities. Each sample is annotated with 1--5 human ratings for
PA and IA, together with detailed rationales under a cross-review protocol.

Evaluation against 29 model-judge baselines and 8 non-MLLM metrics reveals
gaps between existing judges and human judgments. Frontier models show lower
agreement on physical adherence: GPT-5.5~\citep{openai2026gpt55} reaches $r=0.49$
on PA. Specialized video judges reach at most $r=0.42$ on PA and $0.59$ on IA,
while non-MLLM baselines show limited sensitivity to fine-grained physical
correctness. Failure analysis identifies cases in which existing judges overlook
interaction failures and physical hallucinations.

Motivated by these findings, we construct \textbf{\textsc{RoboJudge-Train}},
a large-scale embodied-video judge-training dataset comprising 12K videos
annotated with 25K human-verified scores and model-assisted rationales for PA and IA. We
fine-tune a 9B MLLM from Qwen3.5-9B using supervised learning followed by
reinforcement learning, yielding \textbf{\textsc{RoboJudge-9B}}.
Among the model judges evaluated on \textsc{RoboJudge-Bench}, the resulting
\textsc{RoboJudge-9B} achieves the highest overall agreement, reaching a Pearson
$r$ of 0.719 compared with 0.401 for its base model. Training on
\textsc{RoboJudge-Train} also yields gains on VideoPhy2~\citep{bansal2026videophy} and
Physion-Eval~\citep{zhang2026physion}.
We further use \textsc{RoboJudge-9B} as a downstream evaluator to benchmark
video generation models using the same PA and IA criteria. These
capabilities also suggest broader downstream potential for online physical
rollout evaluation and agentic embodied video generation.
In summary, our contributions are three-fold:

1. \textsc{RoboJudge-Bench}. A benchmark designed for judges of embodied video
   generation: 800 human-annotated rollouts from 8 corpora and 10 video generators,
   spanning grippers, dexterous hands, and human hands in simulated and
   real environments, rated on physical adherence and instruction alignment.

2. A systematic study of existing judges. We compare against 29 model-judge baselines and 8 non-MLLM
   metrics and characterize gaps with human judgments, with failures
   concentrated on fine-grained interaction errors and physical hallucinations.

3. \textsc{RoboJudge-Train} and \textsc{RoboJudge-9B}. We construct a
   training dataset for embodied video judges, comprising 12K
   embodied videos spanning 14 generators and real demonstrations, with 25K human-verified scores and model-assisted rationales,
   together with a 9B judge fine-tuned from
   Qwen3.5-9B via SFT and RL. It achieves the highest overall agreement among
   the evaluated model judges and shows gains on OOD benchmarks.
   As a downstream use case, we apply it to rank video generation models using
   the same PA and IA criteria.

%% file: Sections/related_work.tex
\begin{figure}[t]
    \setlength{\abovecaptionskip}{2pt}
    \centering
    \includegraphics[width=1\linewidth]{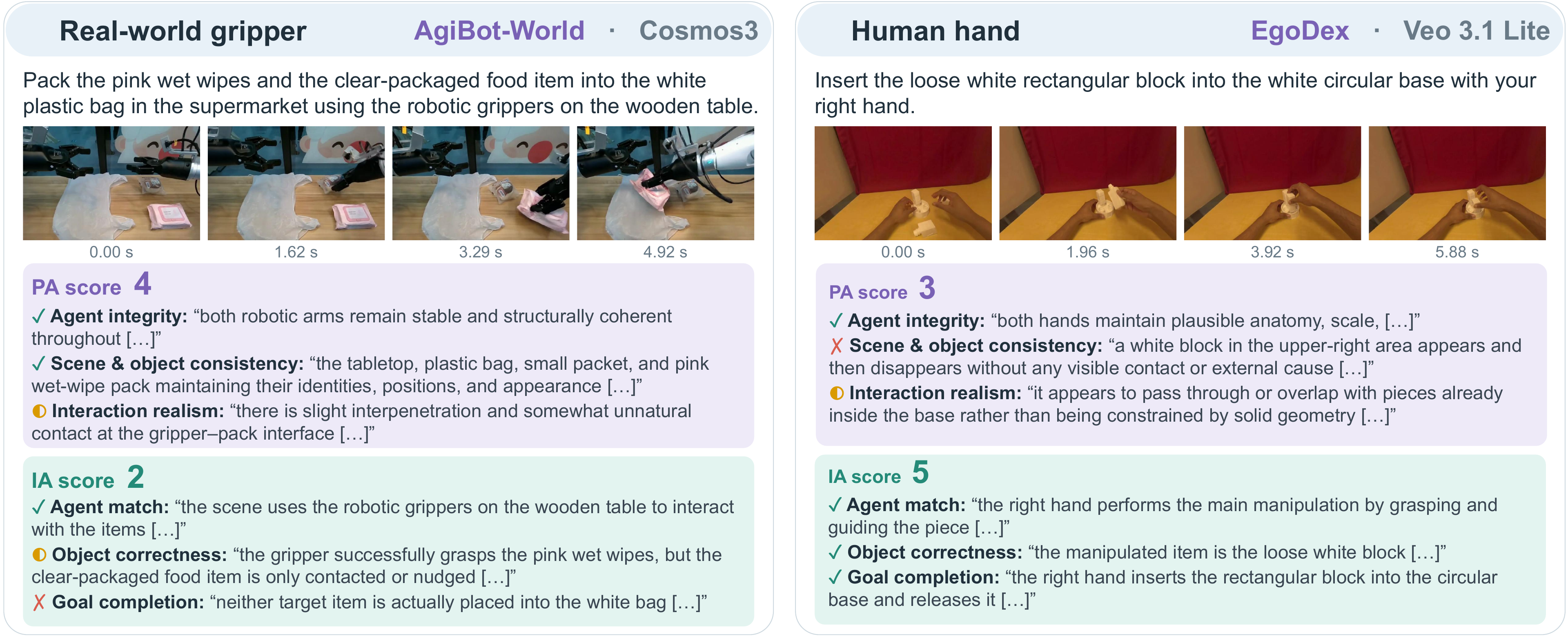}
    \caption{\textbf{Examples from \textsc{RoboJudge-Bench}.} Each example shows the task instruction, sampled video frames, and human ratings with supporting evidence for Physical Adherence (PA) and Instruction Alignment (IA). Left: a gripper-robot video preserves most physical properties but fails to complete the packing task (PA: 4; IA: 2). Right: a human-hand video completes the requested insertion despite object-consistency and interaction artifacts (PA: 3; IA: 5). The examples illustrate why physical plausibility and task completion require separate assessment.}
    \label{fig:example}
\end{figure}

Embodied MLLM benchmarks also diagnose atomic perception, reasoning, and
planning capabilities~\citep{qi2026bear}.

\textbf{Benchmarks for Video Generative Models.} Video-generation
benchmarks have gradually moved beyond perceptual quality toward physical and
embodied evaluation. General benchmarks evaluate visual quality, text-video
alignment, and physical commonsense~\citep{liu2024evalcrafter,
huang2025vbench++,huang2024vbench,videoscore,bansal2025videophy,bansal2026videophy,
meng2024towards,lin2026phyground,qi2026mme}. Recent embodied and world-model
benchmarks further test instruction following, physical adherence, and
task-level behavior in generated rollouts~\citep{worldmodelbench,ewmbench,
fan2026wow,shang2026worldarena,jiang2026robowm}. These benchmarks primarily
measure generated videos or generative models themselves. In contrast, our work
studies whether the \emph{judges} used for embodied video evaluation are
reliable.

\textbf{Video Quality Evaluation Metrics.}
Video evaluation metrics have also expanded in scope. Classical metrics measure
pixel-level or perceptual similarity~\citep{huynh2008scope,bakurov2022structural,
zhang2018unreasonable}, while FVD compares video feature distributions at the
dataset level~\citep{unterthiner2018towards}. Learned representations such as
CLIP, DINO, and V-JEPA provide semantic or temporal features, and optical-flow
estimators such as RAFT support motion-based measures~\citep{radford2021clip,
caron2021dino,bardes2024vjepa,teed2020raft}. These metrics capture useful
signals, but they do not explicitly judge physical validity or task completion.
Recent MLLM-based evaluators add language-conditioned video judgments, including
VideoScore, VideoScore2, VideoPhy2, VideoCon, PhyJudge, and
WBench~\citep{videoscore,videoscore2,bansal2026videophy,bansal2024videocon,
lin2026phyground,ying2026wbench}. However, they are primarily designed for
general-domain video quality, physical plausibility, or world-model evaluation,
rather than joint embodied physical adherence and instruction alignment.

\input{tables/benchmark_comparison}

\textbf{Judge Model Benchmarks.} A parallel line of work studies multimodal
models as judges, with an emphasis on human agreement, reliability, bias, and
calibration~\citep{zheng2023judging,chiang2024chatbotarena,kim2023prometheus,
kim2024prometheus2,mllmasajudge,mjbench,panickssery2024llm}. Existing video
judge benchmarks cover video quality, physical reasoning, preference,
feedback, and artifact judgments, but mostly in general-domain
videos~\citep{videoscore2,bansal2026videophy,liu2025uve,song2025vfeval,
tang2026artifactbench}. Physion-Eval includes egocentric human videos, and
RoboGaze studies generated robot-manipulation videos~\citep{zhang2026physion,
nguyen2026robogaze}. However, they remain limited in embodiment, source, and
generator diversity, as summarized in Table~\ref{tab:related_work_benchmark}.

In contrast, \textsc{RoboJudge-Bench} tests judge reliability rather than generator quality alone, using 800 embodied videos from 8 source corpora and 10 video generators, spanning human hands, dexterous hands, and robot grippers. It provides paired human references for \textit{Physical Adherence} and \textit{Instruction Alignment}. \textsc{RoboJudge-Train} further supplies human-verified dimensional scores and model-assisted rationales for training our dedicated judge, \textsc{RoboJudge-9B}.

%% file: tables/benchmark_comparison.tex
\begin{table*}[t]
\centering
\setlength{\tabcolsep}{3pt}
\caption{Comparison of representative datasets for evaluating MLLMs as world model judges. Numbers in parentheses indicate the number of embodied videos.
}
\label{tab:related_work_benchmark}
\resizebox{0.99\textwidth}{!}{%
\begin{tabular}{l c c c c c l c c c}
\toprule
\textbf{Dataset} &
\textbf{Bench Size} &
\textbf{Training Size} &
\textbf{Domain} &
\textbf{Real} &
\textbf{Sim.} &
\textbf{Embodiments} &
\textbf{Data Sources} &
\textbf{Video Generators} &
\textbf{Embodied Rubrics} \\
\midrule

VideoPhy2
& 3,397 (11)
& 3K
& General
& --
& --
& --
& --
& 7  %
& \xmark
\\

Physion-Eval
& 880 (204)
& 11k
& General
& \cmark
& \xmark
& Human
& 1
& 5
& \xmark
\\

RoboGaze
& 382 (382)
& --
& Embodied
& \cmark
& \cmark
& Dex.
& 3
& 1
& \cmark
\\

\textbf{RoboJudge}
& \textbf{800 (800)}
& \textbf{12K}
& \textbf{Embodied}
& \cmark
& \cmark
& \textbf{Human\&Dex.\&Grip.}
& \textbf{8}
& \textbf{10}
& \cmark

\\
\bottomrule
\end{tabular}%
}
\vspace{1.5mm}
\end{table*}

%% file: Sections/robojudge_bench.tex
\subsection{Overview}
\label{sec:bench_overview}

\begin{figure*}[t]
\centering
\begin{minipage}{\textwidth}
    \subcaptionbox{Key statistics.\label{fig:bench_stats:a}}[0.28\textwidth]{
        \centering
        \scriptsize
        \setlength{\tabcolsep}{3pt}
        \begin{tabular}{@{}lr@{}}
        \toprule
        \textbf{Statistic} & \textbf{Number} \\
        \midrule
        Total test videos        & 800 \\
        \ - gripper robot        & 464 (58.0\%) \\
        \ - dexterous humanoid   & 121 (15.1\%) \\
        \ - human hand           & 215 (26.9\%) \\
        \ - simulated / real     & 131 / 669 \\
        \midrule
        Source corpora           & 8 \\
        Video generators         & 10 \\
        \midrule
        Sub-scores per PA / IA      & 3 \\
        Score scale              & 1--5 \\
        Mean PA / IA             & 3.03 / 3.24 \\
        \midrule
        Max instruction words    & 37 \\
        Avg instruction words    & 17.9 \\
        \bottomrule
        \end{tabular}
    }
    \hfill
    \subcaptionbox{Data distribution.\label{fig:bench_stats:b}}[0.33\textwidth]{
        \centering
        \includegraphics[width=\linewidth]{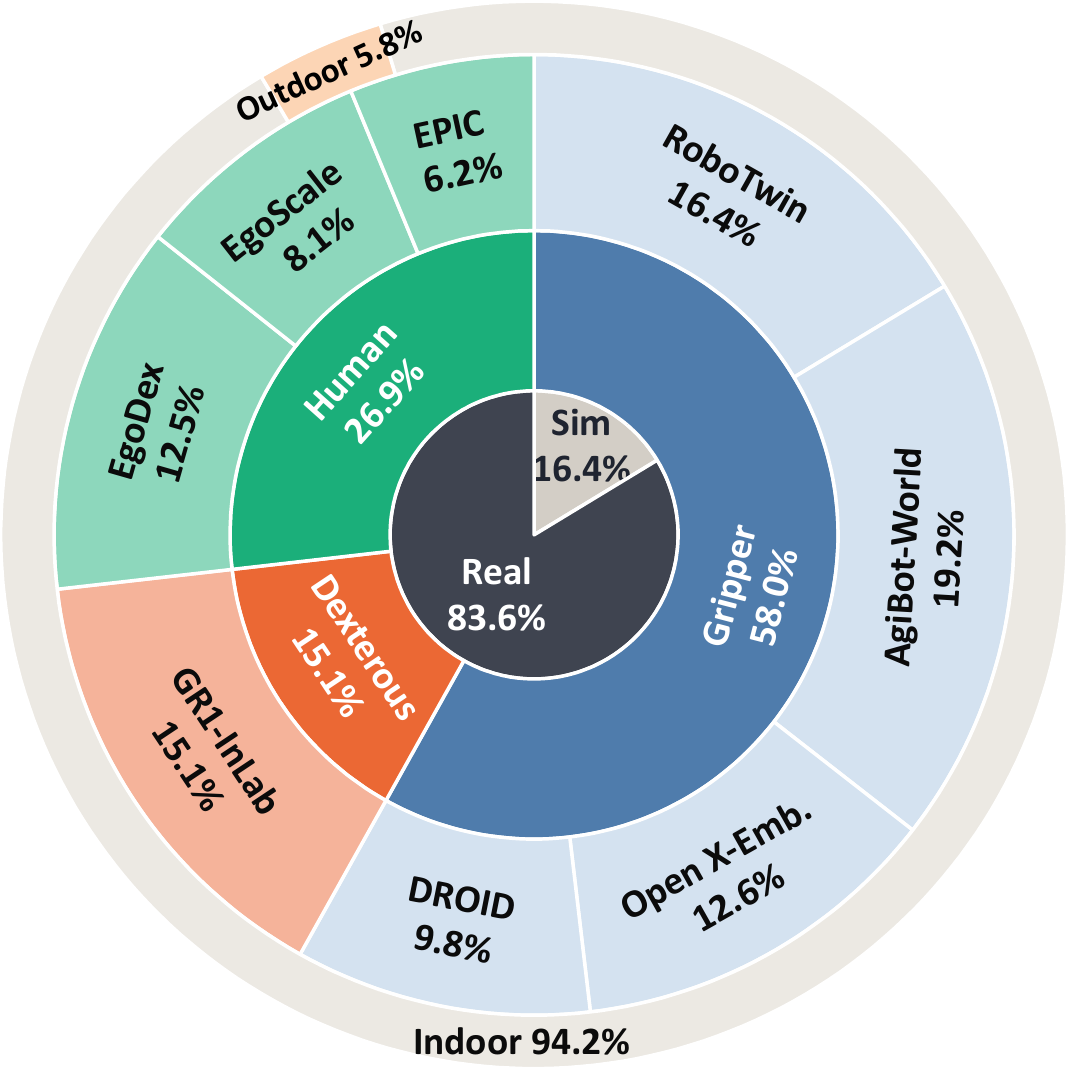}
    }
    \hfill
    \subcaptionbox{Judge model ranking.\label{fig:bench_stats:c}}[0.33\textwidth]{
        \centering
        \includegraphics[width=\linewidth]{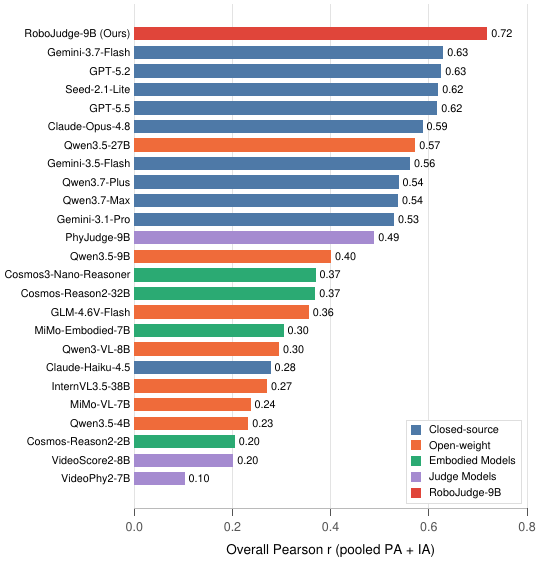}
    }
\end{minipage}
\caption{\textbf{Statistics of \textsc{RoboJudge-Bench}.} (a) Key statistics of the test set.
(b) Videos by embodiment (inner ring) and source corpus (outer ring).
(c) Judge model ranking by pooled Overall Pearson correlation across PA and IA.}
\label{fig:bench_stats}
\end{figure*}

\textsc{RoboJudge-Bench} is a benchmark for evaluating MLLMs as judges of generated embodied-manipulation videos. Rather than evaluating video generators directly, it benchmarks the \emph{judges}: each instance pairs an instruction and a generated video with human annotations along two embodied-specific axes, \textit{Physical Adherence} and \textit{Instruction Alignment}, and judge performance is measured by agreement with these human ratings. Figure~\ref{fig:example} illustrates the two evaluation axes with annotated examples in which physical plausibility and task completion receive different scores. As shown in Figure~\ref{fig:bench_stats}, the benchmark contains 800 generated videos drawn from 8 source corpora, spanning diverse embodiments as well as both real-world and simulated settings. The videos are generated by 10 models with varying capabilities, yielding a broad spectrum of generation quality and human scores across the full rating range.

\subsection{Evaluation Dimensions}
\label{sec:dimensions}

Each judge rates each generated video along two complementary axes, each decomposed into three fine-grained rubric aspects that serve as the primitive units of annotation.
\textit{\textbf{Physical Adherence (PA)}} evaluates whether a video is physically plausible and visually consistent, independent of the instruction, so a rollout that executes the wrong action can still receive full PA marks if its dynamics obey real-world physics. Its aspects proceed outward from the agent: \emph{agent consistency} (the gripper or hand remains structurally coherent and temporally stable), \emph{scene consistency} (the background and objects persist without unrealistic deformation, disappearance, duplication, or sudden displacement), and \emph{interaction realism} (contact, grasping, object motion, and force response follow plausible dynamics).
\textit{\textbf{Instruction Alignment (IA)}} assesses whether the video faithfully executes and completes the manipulation instruction, conditioned on the text and the initial frame, and its aspects follow a verification sequence of who acts, on what, and to what end: \emph{agent match} (the specified agent or end-effector performs the action), \emph{object correctness} (the action targets the correct object with the instructed attributes, such as identity and color), and \emph{goal completion} (the target end-state is fully reached rather than merely approached).
Together, the two axes distinguish physical plausibility from task completion,
which can diverge across current generators.

\subsection{Data Curation}
\label{sec:data_curation}

We curate benchmark source clips from seven real-world embodied-manipulation corpora, including robot-manipulation corpora such as AgiBot-World~\citep{bu2025agibot}, DROID~\citep{droid_2024}, GR1-InLab~\citep{wu2023unleashing}, and Open X-Embodiment~\citep{openx24}, as well as egocentric human-manipulation corpora such as EgoDex~\citep{egodex}, EgoScale~\citep{zheng2026egoscalescalingdexterousmanipulation}, and EPIC-KITCHENS~\citep{Damen2022RESCALING}. We additionally include RoboTwin~\citep{Mu_2025_CVPR} as a simulated manipulation corpus, given its widespread use for robot-policy evaluation. From each clip, we extract an initial-frame--instruction pair. Because raw instructions are often underspecified, we use GPT-5.5 to draft refined instructions and ask human annotators to verify and correct them against the initial frame. Following common practice~\citep{shang2026worldarena,ewmbench}, we derive generation prompts using five prompt-prefix templates and five rewriting templates, yielding ten diverse prompt variants per clip (Appendix~\ref{appendix:video_prompt_engineering}). We use 14 proprietary and open-source video generators to construct the full data pool (Appendix~\ref{appendix:data_statistics}). Before human annotation, we select 800 videos from 10 of these generators for \textsc{RoboJudge-Bench}; \textsc{RoboJudge-Train} additionally includes videos from the remaining four.

We recruit 20 graduate-level annotators, who complete four alignment rounds, each on 100 practice videos, before production labeling; their agreement improves over the course of training (Appendix~\ref{appendix:annotation}, Table~\ref{tab:annotator_alignment}). For the \emph{test set}, which serves as the human reference for evaluating judges, each item is first labeled by one annotator and then independently cross-reviewed by another. An expert adjudicates the two assessments to determine the final ground-truth label. Detailed guidelines and quality-control criteria are provided in Appendix~\ref{appendix:annotation}.
Thus, \textsc{RoboJudge-Bench} uses expert-adjudicated human labels for evaluation, while \textsc{RoboJudge-Train} uses human-verified model-assisted labels for scalable supervision.

%% file: Sections/robojudge_9b.tex
\subsection{Training Data Curation}
\label{sec:training_data}

We construct \textsc{RoboJudge-Train} using the same video-generation
pipeline as \textsc{RoboJudge-Bench}, while avoiding data
leakage by using source clips that are disjoint from those used to
construct the benchmark. We expand the source pool with additional corpora, including LIBERO~\citep{liu2023libero} and DreamDojo-HV~\citep{gao2026dreamdojo}, resulting in 12,351 training videos: 11,520 generated videos from 14 generators and 831 verified successful real demonstrations paired with their corresponding instructions.
Each
video receives separate annotations along the two evaluation axes,
Physical Adherence (PA) and Instruction Alignment (IA), yielding 24,702
axis-level training samples in total.

We annotate the expanded training set with the same human annotator pool
using a human-in-the-loop labeling pipeline. Specifically, three
closed-source MLLMs (Gemini-3.5-Flash, GPT-5.5, and Seed-2.1-Lite) first
provide candidate PA and IA scores. Their predictions are averaged to
produce draft scores, which are subsequently reviewed and revised by human
annotators. Annotators also provide brief evidence notes to justify their
final scores. Finally, GPT-5.5 rewrites the human-authored evidence notes into
structured reasoning traces conditioned on the human-verified scores. These
annotations and model-assisted reasoning traces are used as supervision for
training \textsc{RoboJudge-9B}.
\subsection{Model Training}
\label{sec:model_training}

We initialize \textsc{RoboJudge-9B} from \textbf{Qwen3.5-9B} and train it
with a two-stage strategy consisting of supervised fine-tuning (SFT)
followed by reinforcement learning (RL).

\textbf{Supervised Fine-Tuning.}
We first perform full-parameter supervised fine-tuning on
\textsc{RoboJudge-Train}, starting from the pretrained Qwen3.5-9B
checkpoint. We train for one epoch and use the resulting SFT checkpoint
to initialize the subsequent RL stage.

\textbf{Reinforcement Learning.}
We further optimize the SFT model with Group Relative Policy Optimization
(GRPO). We directly optimize the exact-score agreement objective. For each
generated response, the reward is defined as
\begin{equation}
R_{\mathrm{acc}} =
\begin{cases}
1, & \text{if the predicted score exactly matches the reference},\\
0, & \text{otherwise}. \\
\end{cases}
\end{equation}
Starting from the
SFT checkpoint, we train for 200 RL steps. The resulting checkpoint is
used as \textsc{RoboJudge-9B}. We provide training details in Appendix~\ref{appendix:exp_setup}.

%% file: Sections/experiments.tex
\subsection{Evaluated Models and Baselines}
\label{sec:evaluated_models}

We compare \textsc{RoboJudge-9B} against 29 model-judge baselines on \textsc{RoboJudge-Bench}, including \textbf{10 closed-source MLLMs}, \textbf{7 open-weight MLLMs}, \textbf{7 specialized embodied MLLMs}, and \textbf{5 specialized judge models}. General-purpose and embodied MLLMs use the unified \textsc{RoboJudge-Bench} prompt, while specialized judge models use their original prompts. We additionally include \textbf{8 non-MLLM metrics}, following metric families from VBench~\citep{huang2024vbench}, EvalCrafter~\citep{liu2024evalcrafter}, and WorldArena~\citep{shang2026worldarena}. Full evaluation protocols are provided in Appendix~\ref{appendix:eval_protocol}.

\subsection{Evaluation Metrics}
\label{sec:metrics}

\begin{figure}[t]
    \centering
    \includegraphics[width=1\linewidth]{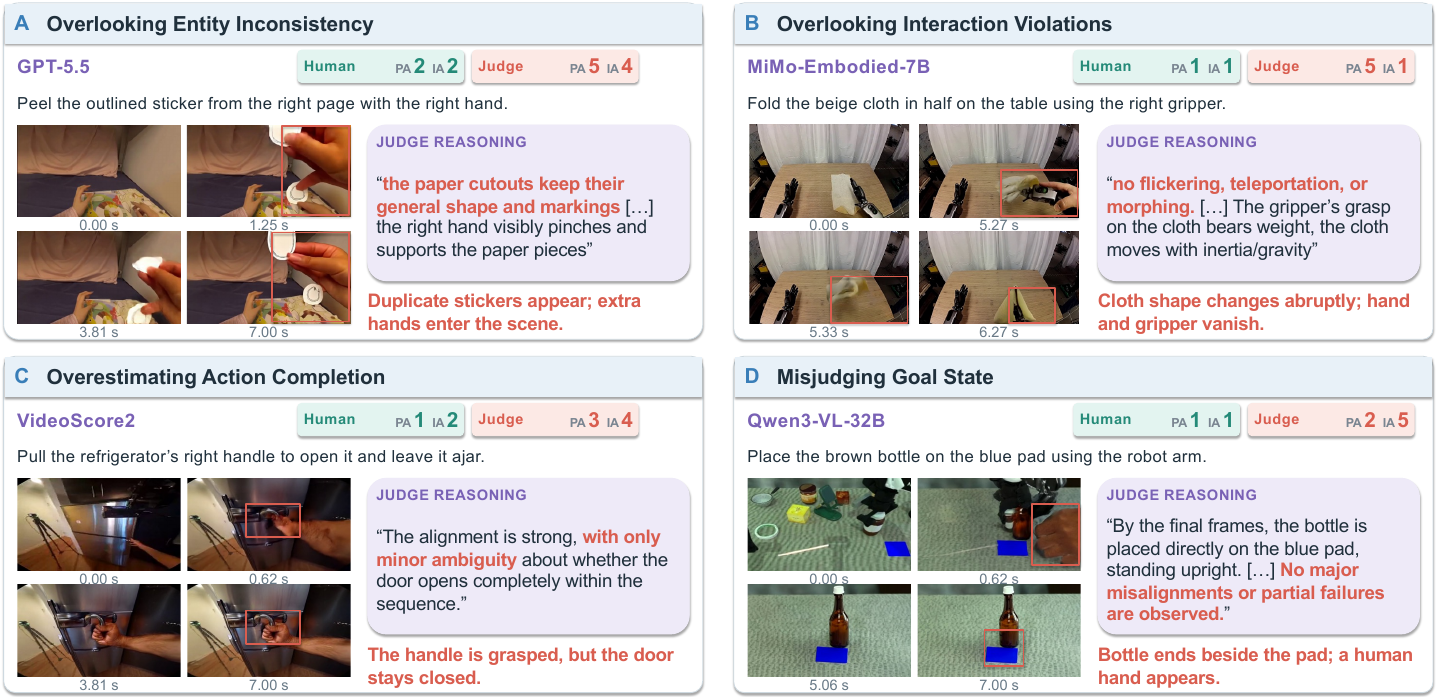}
    \caption{\textbf{Qualitative failure cases of existing judges on \textsc{RoboJudge-Bench}.} Each panel compares human and judge Physical Adherence (PA) and Instruction Alignment (IA) scores with sampled frames and judge reasoning. (A) GPT-5.5 misses duplicated stickers and extra hands. (B) MiMo-Embodied-7B misses abrupt cloth deformation and vanishing manipulators. (C) VideoScore2 credits task completion although the refrigerator door stays closed. (D) Qwen3-VL-32B reports successful placement although the bottle lands beside the target pad. These failures span entity consistency, interaction realism, action completion, and goal-state verification.}
    \label{fig:failure}
\end{figure}

After collecting judgments from each model, we quantify alignment with
human annotations along the two axes of our benchmark using three metrics. \textbf{(1) Pearson correlation.} Following LLM-as-a-Judge~\citep{zheng2023judging}
and MLLM-as-a-Judge~\citep{mllmasajudge}, we take the Pearson
correlation~\citep{pearsoncorr} between model scores and human ratings as our
primary metric. \textbf{(2) Exact-match accuracy (Acc).} We report agreement with
human ratings at the item level by measuring the fraction of
items whose predicted score equals the human score. \textbf{(3) Mean squared error.} We report the mean squared error (MSE) between
predicted and human scores, which, unlike correlation-based metrics, is sensitive to a
systematic shift in a judge's scoring scale.

\subsection{Results and Analysis}
\label{sec:overall_judge}

\input{tables/judge_performance}

\input{tables/non_mllm_comparison}

Table~\ref{tab:judge_performance} shows that existing judges struggle most with physical adherence and that non-MLLM metrics and specialized video judges align poorly with human judgments on embodied videos. Among model judges, \textbf{RoboJudge-9B performs best}, reaching $r=0.719$ overall, with $r=0.645$ on PA and $r=0.775$ on IA; paired cluster-bootstrap analysis confirms robust gains in IA and overall agreement (Appendix~\ref{appendix:bootstrap}). We compare conventional non-MLLM metrics in Figure~\ref{fig:non_MLLM_baseline}, while Figure~\ref{fig:judge_landscape} in Appendix~\ref{appendix:judge_landscape} plots correlation against MSE.

\paragraph{Non-MLLM metrics remain weak proxies for physical judgment.}
As shown in Figure~\ref{fig:non_MLLM_baseline}, non-MLLM metrics correlate only moderately with human PA ratings: MUSIQ and pixel consistency are weak ($r=0.078$ and $0.155$), while DINO and V-JEPA Paired Similarity reach $r=0.392$ and $0.370$. Both remain below \textsc{RoboJudge-9B}, showing that visual quality and representation similarity are insufficient for embodied physical plausibility.

\paragraph{Existing judges struggle more with physical adherence.}
Before task-specific training, the best IA correlation ($r=0.733$) is well above the best PA correlation ($r=0.611$), and specialized embodied MLLMs remain at $r \le 0.261$ on PA. Figure~\ref{fig:failure} shows why: judges often miss entity and interaction inconsistencies (A--B), or infer task completion from incorrect final states (C--D).

\paragraph{Existing embodied VLMs and specialized judges do not transfer well to
embodied video evaluation.}
Existing embodied MLLMs and specialized video judges show limited agreement with human ratings. RoboBrain2.5-8B and RynnBrain-1.1-9B are near random on PA ($r \le 0.07$), while VideoPhy2, VideoScore2, and VideoCon reach at most $r=0.20$. PhyJudge is stronger ($r=0.417$) but still trails RoboJudge-9B ($r=0.645$).

\paragraph{RoboJudge reduces directional scoring bias.}
Beyond correlation, several baselines show clear directional bias, while \textsc{RoboJudge-9B} is more balanced on both PA and IA, suggesting better calibration to human scores (Appendix~\ref{appendix:additional_analysis}).

\subsection{Ablation Study}
\label{sec:ablation}

\textbf{Frame Rate.}
We ablate the video sampling rate at 2, 4, and 8 FPS for GPT-5.5 and
Qwen3.5-9B. GPT-5.5 shows no monotonic benefit from denser sampling: its
overall correlation is 0.620, 0.618, and 0.620 at 2, 4, and 8 FPS,
respectively. Qwen3.5-9B improves from 0.363 at 2 FPS to 0.401 at 4 FPS,
while 8 FPS yields no further gain. We therefore use \textbf{4 FPS} as the
common default setting, which captures the improvement of the open-weight
model without the additional visual-token cost of 8 FPS. By construction, the
4 FPS rows reproduce the corresponding main-table results (full results in
Table~\ref{tab:fps_ablation}, Appendix~\ref{appendix:fps_ablation}).

\textbf{Training Stages.} We further ablate the contributions of supervised fine-tuning (SFT) and reinforcement learning (RL). SFT substantially improves the base Qwen3.5-9B on both PA and IA, demonstrating the effectiveness of \textsc{RoboJudge-Train} for task-specific supervision. Building on SFT, RL further improves the overall correlation and Acc, increasing the overall $r$ from 0.700 to 0.719. We therefore use the \textbf{SFT+RL} model as our final \textsc{RoboJudge-9B}; full results are reported in Table~\ref{tab:qwen35_9b_sft_rl_ablation}.
\label{appendix:stage_ablation}
\input{tables/training_stage_ablation}

\subsection{Generalization Study}
\label{sec:ood_generalization}

\input{tables/ood_evaluation}
As shown in Table~\ref{tab:ood_evaluation}, training on
\textsc{RoboJudge-Train} substantially improves the judge's performance on
both external OOD benchmarks, despite the distribution shift and differences
in evaluation prompts. On VideoPhy2, both SFT and SFT+RL substantially improve
over the Qwen3.5-9B base model, with the SFT+RL model reaching Pearson
correlations of 0.346 and 0.572 on PC and SA, respectively. On Physion-Eval, the SFT model
achieves a Youden's $J$ of 0.105, substantially outperforming the base
model (0.056). These results demonstrate that the
judgment capability learned from \textsc{RoboJudge-Train} transfers beyond
the training distribution and remains effective under substantial
distribution shifts.

\subsection{Downstream Application: Video Generation Leaderboard}
\label{sec:videogen_leaderboard}

\input{Sections/downstream_applications}

%% file: tables/judge_performance.tex
\begin{table}[t]
\setlength{\belowcaptionskip}{2pt}
\centering
\caption{Judge performance on \textsc{RoboJudge-Bench}. $r$ is Pearson correlation with human ratings, and Acc is exact-match accuracy. Overall $r$ and MSE pool PA and IA as axis-level score predictions, while Overall Acc is the average of PA and IA Acc. The best value in each column is shown in \textbf{bold}.}
\label{tab:judge_performance}
\fontsize{7.2pt}{9pt}\selectfont
\setlength{\tabcolsep}{3pt}
\begin{tabular}{@{}l ccc ccc cc c@{}}
\hline
 & \multicolumn{3}{c}{\textbf{Physical Adherence}} & \multicolumn{3}{c}{\textbf{Instruction Alignment}} & \multicolumn{3}{c}{\textbf{Overall}} \\
\cmidrule(lr){2-4} \cmidrule(lr){5-7} \cmidrule(lr){8-10}
\textbf{Judge} & $r\uparrow$ & Acc$\uparrow$ & MSE$\downarrow$ & $r\uparrow$ & Acc$\uparrow$ & MSE$\downarrow$ & $r\uparrow$ & Acc$\uparrow$ & MSE$\downarrow$ \\

\midrule
Random Baseline
& 0.000 & 0.200 & 3.349
& 0.000 & 0.200 & 3.631
& 0.000 & 0.200 & 3.490 \\
\midrule
\multicolumn{10}{c}{\textit{\textbf{Closed-Source MLLMs}}}\\
\midrule
Gemini-3.7-Flash~\citep{google2026gemini37flash}
& 0.611 & 0.245 & 1.949
& 0.709 & 0.341 & 0.916
& 0.630 & 0.293 & 1.433 \\

Gemini-3.5-Flash~\citep{google2026gemini35flash}
& 0.594 & 0.304 & 1.527
& 0.535 & 0.384 & 1.400
& 0.562 & 0.344 & 1.463 \\

Gemini-3.1-Pro~\citep{google2026gemini31pro}
& 0.451 & 0.296 & 1.505
& 0.584 & 0.321 & 1.395
& 0.529 & 0.309 & 1.450 \\

GPT-5.5~\citep{openai2026gpt55}
& 0.489 & 0.334 & 1.359
& 0.733 & 0.481 & 0.786
& 0.618 & 0.407 & 1.073 \\

GPT-5.2~\citep{openai2025gpt52}
& 0.499 & 0.366 & 1.159
& 0.722 & 0.477 & 0.851
& 0.626 & 0.422 & 1.005 \\

Claude-Opus-4.8~\citep{anthropic2026claudeopus48}
& 0.505 & 0.339 & 1.282
& 0.655 & 0.340 & 1.468
& 0.587 & 0.339 & 1.375 \\

Claude-Haiku-4.5~\citep{anthropic2025claudehaiku45}
& 0.179 & 0.246 & 2.639
& 0.357 & 0.333 & 2.442
& 0.278 & 0.289 & 2.541 \\

Qwen3.7-Max~\citep{qwen2026qwen37max}
& 0.553 & 0.239 & 2.466
& 0.539 & 0.365 & 2.055
& 0.537 & 0.302 & 2.261 \\

Qwen3.7-Plus~\citep{qwen2026qwen37plus}
& 0.548 & 0.244 & 2.450
& 0.546 & 0.375 & 2.016
& 0.538 & 0.309 & 2.233 \\

Seed-2.1-Lite~\citep{seed2026seed21}
& 0.590 & 0.385 & 0.935
& 0.694 & 0.445 & 1.120
& 0.619 & 0.415 & 1.028 \\

\midrule
\multicolumn{10}{c}{\textit{\textbf{Open-weight MLLMs}}}\\
\midrule
Qwen3.5-27B~\citep{qwen2026qwen35}
& 0.431 & 0.303 & 1.664
& 0.676 & 0.375 & 1.345
& 0.573 & 0.339 & 1.505 \\

Qwen3.5-9B~\citep{qwen2026qwen35}
& 0.379 & 0.223 & 3.011
& 0.461 & 0.347 & 2.314
& 0.401 & 0.285 & 2.663 \\

Qwen3.5-4B~\citep{qwen2026qwen35}
& 0.063 & 0.154 & 4.513
& 0.383 & 0.331 & 2.620
& 0.232 & 0.242 & 3.566 \\

Qwen3-VL-8B~\citep{bai2025qwen3vl}
& 0.292 & 0.161 & 3.776
& 0.341 & 0.279 & 3.094
& 0.295 & 0.220 & 3.435 \\

GLM-4.6V-Flash~\citep{vteam2025glm45vglm41vthinkingversatilemultimodal}
& 0.253 & 0.163 & 3.845
& 0.440 & 0.289 & 3.058
& 0.356 & 0.226 & 3.451 \\

InternVL3.5-38B~\citep{wang2025internvl3_5}
& 0.204 & 0.127 & 4.691
& 0.375 & 0.308 & 3.044
& 0.271 & 0.218 & 3.867 \\

MiMo-VL-7B~\citep{coreteam2025mimovltechnicalreport}
& 0.069 & 0.138 & 4.710
& 0.362 & 0.262 & 3.412
& 0.237 & 0.200 & 4.061 \\

\midrule
\multicolumn{10}{c}{\textit{\textbf{Specialized Embodied MLLMs}}}\\
\midrule
Cosmos3-Nano-Reasoner~\citep{nvidia2026cosmos3}
& 0.256 & 0.155 & 4.244
& 0.524 & 0.381 & 2.049
& 0.369 & 0.268 & 3.146 \\

Cosmos-Reason2-32B~\citep{nvidia2025cosmosreason2}
& 0.261 & 0.161 & 3.735
& 0.484 & 0.364 & 2.330
& 0.369 & 0.263 & 3.033 \\

Cosmos-Reason2-2B~\citep{nvidia2025cosmosreason2}
& 0.085 & 0.125 & 4.895
& 0.309 & 0.245 & 3.586
& 0.205 & 0.185 & 4.241 \\

MiMo-Embodied-7B~\citep{hao2025mimoembodied}
& 0.209 & 0.119 & 4.736
& 0.447 & 0.300 & 2.807
& 0.304 & 0.209 & 3.776 \\

RynnBrain-1.1-9B~\citep{damo2026rynnbrain11}
& 0.051 & 0.115 & 5.077
& 0.196 & 0.251 & 4.025
& 0.129 & 0.183 & 4.551 \\

RoboBrain2.5-8B~\citep{tan2026robobrain25}
& 0.042 & 0.128 & 4.952
& 0.197 & 0.259 & 3.761
& 0.122 & 0.193 & 4.357 \\

Orca-4B~\citep{orca2026}
& 0.108 & 0.145 & 4.447
& 0.096 & 0.244 & 3.242
& 0.059 & 0.194 & 3.845 \\

\midrule
\multicolumn{10}{c}{\textit{\textbf{Specialized Judge Models}}}\\
\midrule
VideoScore2-8B~\citep{videoscore2}
& 0.195 & 0.311 & 1.455
& 0.191 & 0.255 & 1.725
& 0.202 & 0.283 & 1.590 \\

VideoPhy2-7B~\citep{bansal2026videophy}
& 0.152 & 0.290 & 1.866
& 0.147 & 0.291 & 1.732
& 0.103 & 0.291 & 1.799 \\

VideoCon-7B~\citep{bansal2024videocon}
& 0.095 & 0.150 & 4.079
& 0.055 & 0.164 & 4.633
& 0.077 & 0.157 & 4.356 \\

WBench-30B-A3B~\citep{ying2026wbench}
& 0.412 & 0.319 & 1.197
& -- & -- & --
& -- & -- & -- \\

PhyJudge-9B~\citep{lin2026phyground}
& 0.417 & 0.314 & 1.429
& 0.585 & 0.376 & 1.072
& 0.489 & 0.345 & 1.252 \\

\textbf{RoboJudge-9B (ours)}
& \textbf{0.645} & \textbf{0.434} & \textbf{0.806}
& \textbf{0.775} & \textbf{0.512} & \textbf{0.688}
& \textbf{0.719} & \textbf{0.473} & \textbf{0.747} \\

\hline
\end{tabular}
\end{table}

%% file: tables/non_mllm_comparison.tex
\begin{figure}[t]
\centering
\begin{minipage}[b]{0.385\linewidth}
\centering
\includegraphics[width=\linewidth]{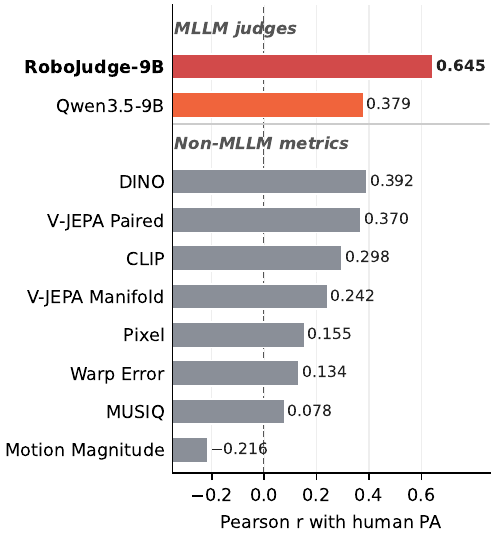}
\caption{Pearson correlation ($r$) with human-rated Physical Adherence (PA) for non-MLLM similarity metrics and MLLM-as-judges.}
\label{fig:non_MLLM_baseline}
\end{minipage}\hfill
\begin{minipage}[b]{0.595\linewidth}
\centering
\includegraphics[width=\linewidth]{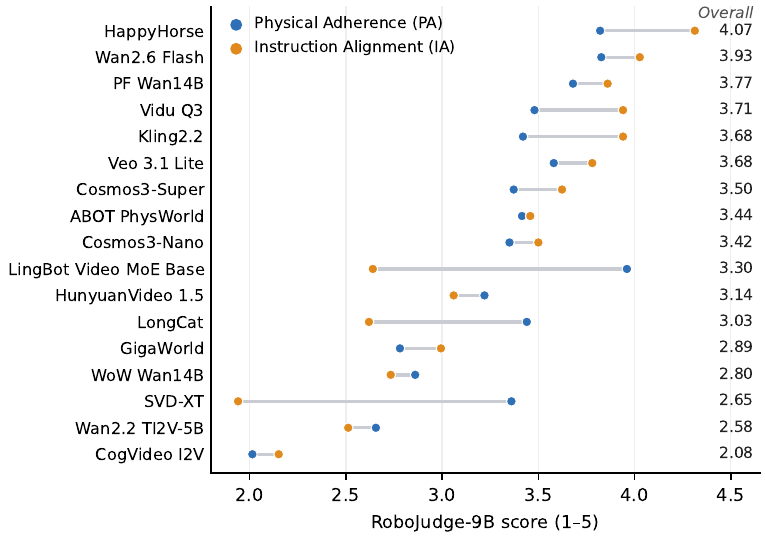}
\caption{Downstream application: ranking video generation models with \textsc{RoboJudge-9B}. Models are sorted by Overall (mean of PA and IA); full scores are in Table~\ref{tab:videogen_leaderboard}.}
\label{fig:videogen_leaderboard}
\end{minipage}
\end{figure}

%% file: tables/training_stage_ablation.tex
\begin{table}[t]
\setlength{\abovecaptionskip}{2pt}
\centering
\caption{Training-stage ablation of \textsc{RoboJudge-9B}. Acc is exact-match accuracy.}
\label{tab:qwen35_9b_sft_rl_ablation}
\fontsize{7.5pt}{8.5pt}\selectfont
\setlength{\tabcolsep}{8pt}
\begin{tabular}{@{}l ccc ccc ccc@{}}
\toprule
\textbf{Model} &
\multicolumn{3}{c}{\textbf{PA}} &
\multicolumn{3}{c}{\textbf{IA}} &
\multicolumn{3}{c}{\textbf{Overall}} \\
\cmidrule(lr){2-4}
\cmidrule(lr){5-7}
\cmidrule(lr){8-10}
& $r\uparrow$ & Acc$\uparrow$ & MSE$\downarrow$
& $r\uparrow$ & Acc$\uparrow$ & MSE$\downarrow$
& $r\uparrow$ & Acc$\uparrow$ & MSE$\downarrow$ \\
\midrule

Qwen3.5-9B
& 0.379 & 0.223 & 3.011
& 0.461 & 0.347 & 2.314
& 0.401 & 0.285 & 2.663 \\

+ SFT
& 0.624 & 0.405 & 0.913
& 0.766 & 0.471 & 0.671
& 0.700 & 0.438 & 0.792 \\

+ SFT + RL (RoboJudge-9B)
& \textbf{0.645} & \textbf{0.434} & \textbf{0.806}
& \textbf{0.775} & \textbf{0.512} & 0.688
& \textbf{0.719} & \textbf{0.473} & \textbf{0.747} \\

\bottomrule
\vspace{-5mm}
\end{tabular}
\end{table}

%% file: tables/ood_evaluation.tex
\begin{wraptable}{r}{0.5\linewidth}
\setlength{\abovecaptionskip}{2pt}
\centering
\vspace{-4mm}
\caption{\textbf{OOD evaluation on VideoPhy2 and Physion-Eval.}
Training on \textsc{RoboJudge-Train} improves performance across all
evaluated OOD metrics over the Qwen3.5-9B base model. $J$ is Youden's $J$.}
\label{tab:ood_evaluation}
\footnotesize
\setlength{\tabcolsep}{3pt}
\begin{tabular}{@{}l cc c@{}}
\toprule
& \multicolumn{2}{c}{\textbf{VideoPhy2}}
& \textbf{Physion-Eval} \\
\cmidrule(lr){2-3}
\textbf{Model}
& \textbf{PC $r$}$\uparrow$
& \textbf{SA $r$}$\uparrow$
& \textbf{$J$}$\uparrow$ \\
\midrule
Qwen3.5-9B
& 0.179 & 0.412 & 0.056 \\

+ SFT
& 0.299 & 0.552 & \textbf{0.105} \\

+ SFT + RL (Ours)
& \textbf{0.346} & \textbf{0.572} & 0.093 \\

\bottomrule
\end{tabular}
\vspace{-3pt}
\end{wraptable}

%% file: Sections/downstream_applications.tex
Beyond measuring judge quality, a human-aligned judge enables a practical downstream use: benchmarking embodied video generation models without collecting new human ratings.
We apply \textsc{RoboJudge-9B} with the same PA and IA rubric to 17 video generation models on a shared set of 50 prompts (Figure~\ref{fig:videogen_leaderboard}; full scores in Table~\ref{tab:videogen_leaderboard}, Appendix~\ref{appendix:videogen_leaderboard}).
PA and IA reveal distinct strengths across generators. HappyHorse and Wan2.6 Flash score highly on both, whereas LingBot Video MoE Base, LongCat, and SVD-XT receive relatively high PA but much lower IA: they produce physically plausible motion while failing the instructed task.
A single overall score would hide this difference, so embodied video generation should be evaluated along both PA and IA dimensions.

Beyond offline benchmarking, \textsc{RoboJudge-9B}'s human-aligned scores may
guide online physical rollout selection in embodied planning (e.g.,
Best-of-$N$). The model may further support quality assessment in agentic
embodied video generation.

%% file: Sections/conclusion.tex
We introduced \textsc{RoboJudge-Bench}, a benchmark for
evaluating whether MLLMs can judge embodied video generation along physical
adherence and instruction alignment. Our study shows that existing judges,
including frontier MLLMs and specialized video evaluators, remain poorly
aligned with human judgments on embodied physical interactions. To address this gap, we
constructed \textsc{RoboJudge-Train} and trained \textsc{RoboJudge-9B}, which
achieves the strongest overall agreement among model judges on
\textsc{RoboJudge-Bench} and transfers to external physical-reasoning benchmarks.
These results highlight the value of human-calibrated PA and IA supervision for
embodied video evaluation.
Together, these capabilities suggest broader downstream potential across
online physical rollout evaluation and agentic embodied video generation.

%% file: appendix/annotation.tex
\section{Annotation Protocol and Interface}
\label{appendix:annotation}

\begin{figure}[h]
\centering
\vspace{-1mm}
\includegraphics[width=0.85\linewidth]{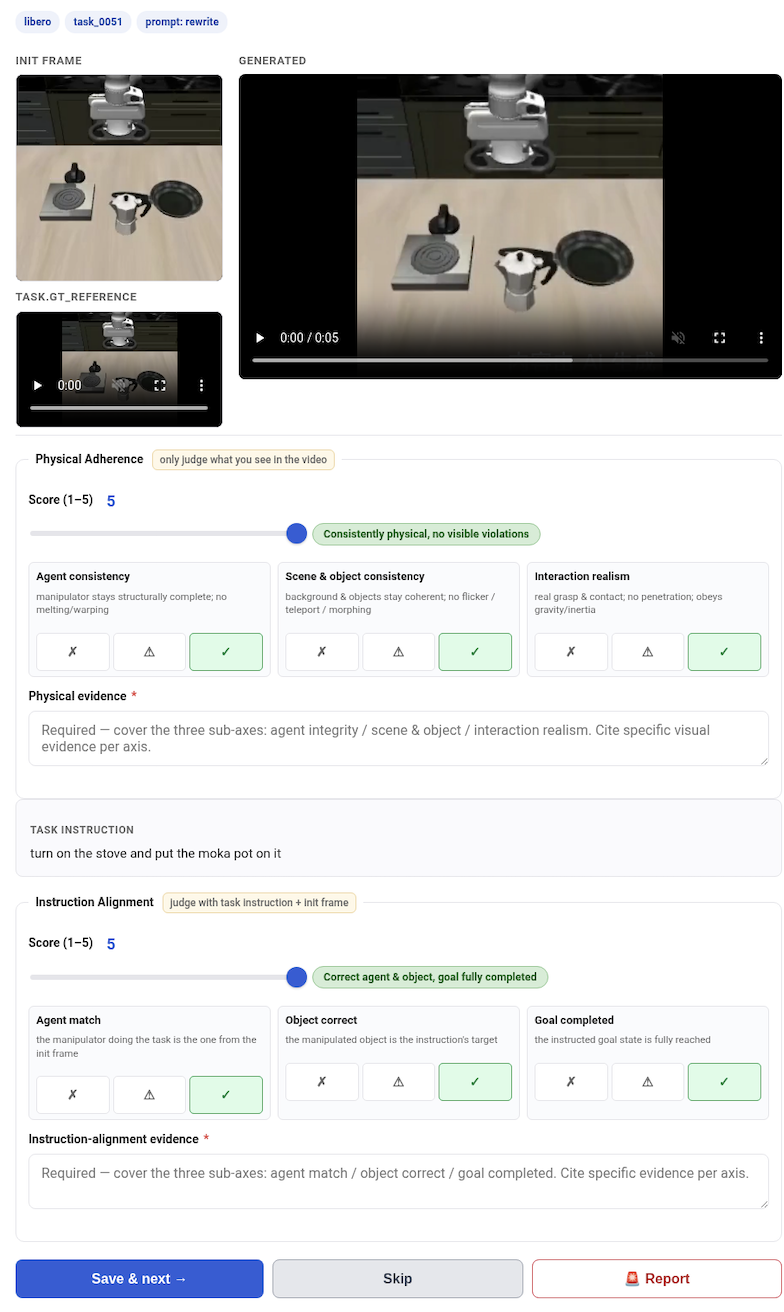}
\vspace{-3mm}
\caption{\textbf{Annotation interface.} Annotators score each generated
video on physical adherence and instruction alignment, complete the corresponding
diagnostic checklists, and write evidence for their decisions.}
\label{fig:annotation_ui}
\vspace{-3mm}
\end{figure}

\subsection{Annotator Training and Calibration}
Before production annotation began, annotators completed four
alignment rounds, each based on a pool of 100 practice videos, for both physical
adherence (PA) and instruction alignment (IA). Each round used the same
checklist as the main annotation: annotators assigned a $1$--$5$ axis score,
filled the three $0$--$2$ diagnostic checklist items for that axis, and wrote
short evidence comments. PA judgments were made
without access to the task instruction, while IA judgments were conditioned on the
instruction and initial frame. Annotation reviewers compared the submissions against the
rubric definitions in Sec.~\ref{sec:dimensions}, gave feedback on disagreements,
and used later rounds to verify that annotators applied the score scale and
diagnostic checklist consistently.

\begin{table}[h]
\centering
\caption{Annotator alignment across four training rounds, each based on 100
practice videos. Metrics aggregate pairwise comparisons of annotators' primary
PA and IA scores.}
\label{tab:annotator_alignment}
\small
\setlength{\tabcolsep}{9pt}
\begin{tabular}{l r r r}
\toprule
\textbf{Round} & \textbf{Exact} & \textbf{Within $\boldsymbol{\pm 1}$} & \textbf{Mean abs. diff.} \\
\midrule
1 & 32.8\% & 76.2\% & 0.98 \\
2 & 39.0\% & 78.1\% & 0.97 \\
3 & 38.3\% & 81.2\% & 0.85 \\
4 & 52.9\% & 92.1\% & 0.58 \\
\bottomrule
\end{tabular}
\end{table}

As shown in Table~\ref{tab:annotator_alignment}, exact agreement increases from
32.8\% to 52.9\%, while agreement within one point rises from 76.2\% to
92.1\%; the mean absolute difference correspondingly decreases from 0.98 to
0.58. These trends indicate progressively tighter application of the scoring
rubric before production annotation. For benchmark ground-truth construction,
one annotator first assigns each item's scores and supporting evidence, and a
second annotator independently cross-reviews the complete annotation. An expert
then adjudicates the two assessments and determines the final label.

\subsection{Model-Assisted Training Annotation}
The training split was annotated with a model-assisted workflow. Three
closed-source MLLMs (Gemini-3.5-Flash, GPT-5.5, and Seed-2.1-Lite) first generated candidate PA/IA scores and checklist
labels. We averaged their scores to form
an initial label proposal. Human annotators then reviewed the proposed labels in
the annotation interface, corrected numeric scores or checklist labels when the
pre-annotation missed or overstated a violation, and wrote brief evidence notes
citing the relevant visual evidence. GPT-5.5 then rewrote these human-authored notes into structured reasoning traces conditioned on the human-verified scores (Sec.~\ref{sec:training_data}). This workflow reduced the cost of
covering the large training pool, while the supervision used for \textsc{RoboJudge-9B}
reflected human revisions rather than raw model outputs.

\subsection{Annotation Interface}
Figure~\ref{fig:annotation_ui} shows the basic annotation page used by human
annotators. The page presents the generated video and reference context, collects
the PA and IA axis scores, asks for the six diagnostic checklist labels, and
records free-text evidence for the annotator's decision. Model identity is
hidden from annotators during labeling to reduce generator-specific bias.

%% file: appendix/evaluation_protocol.tex
\section{Evaluation Protocol}
\label{appendix:eval_protocol}

This appendix specifies the evaluation protocols used for all model judges and
non-MLLM metric baselines. We separate the protocol into three parts: the
unified prompt and video-input setting for general and embodied VLMs, the
original protocols used by specialized VLM judges, and the implementation of
non-MLLM metrics.

\input{appendix/judge_prompts}

\subsection{Specialized VLM Judges}
\label{appendix:specialized_vlm_judges}

Specialized VLM judges are evaluated on the same \textsc{RoboJudge-Bench} test
set, but unlike general and embodied VLMs, they are run with their original
model-specific scoring interfaces and prompts whenever those interfaces are
available. This preserves the intended usage of each judge and avoids forcing
specialized evaluators into the unified JSON prompt in Appendix~\ref{appendix:judge_prompt}.
We evaluate the following specialized VLM judges.

\begin{itemize}[leftmargin=*]
    \item \textbf{VideoScore2}~\citep{videoscore2} uses a deliberative
    video-scoring procedure. We use its native $1$--$5$ physical/common-sense
    consistency and text-to-video alignment scores as PA and IA, respectively.
    \item \textbf{VideoPhy2}~\citep{bansal2026videophy} focuses on physical
    commonsense and temporal consistency. We use its native $1$--$5$ physical
    commonsense and semantic-adherence scores as PA and IA, respectively.
    \item \textbf{VideoCon}~\citep{bansal2024videocon} evaluates video-language
    consistency through contrastive scoring. We use its physical-commonsense
    score as PA and semantic-adherence score as IA, mapping each native score
    $s\in[0,1]$ to $1+\operatorname{round}(4s)$.
    \item \textbf{PhyJudge}~\citep{lin2026phyground} evaluates physical
    reasoning in generative world-model videos. We use the mean of its physical
    trustworthiness and persistence scores as PA, and its semantic-adherence
    score as IA.
    \item \textbf{WBench}~\citep{ying2026wbench} provides a visual-quality
    evaluator for interactive video world models. We map its visual-quality
    score to Physical Adherence (PA). Because WBench does
    not evaluate instruction following, we do not report Instruction Alignment
    (IA) or overall results for this model.
\end{itemize}

\input{appendix/non_mllm_metrics}

%% file: appendix/judge_prompts.tex
\subsection{General and Embodied VLMs}
\label{appendix:judge_prompt}

General-purpose VLMs and embodied VLMs are evaluated under a unified
\textsc{RoboJudge-Bench} setting. For each generated video, the runner provides
the model with a structured video input sampled at 4 fps with at most 32
uniformly sampled frames. All models in this group receive the same prompt and
are required to return the same JSON fields for Physical Adherence and
Instruction Alignment, ensuring that differences in scores reflect judge
behavior rather than prompt or output-format changes.

\subsubsection{Physical Adherence}

\paragraph{System prompt.}
\begin{Verbatim}[breaklines,breaksymbolleft={}]
You are a strict, calibrated evaluator of the PHYSICAL REALISM of
AI-generated embodied / robot-manipulation videos (a robot arm/gripper or
a human hand acting on objects). You are shown uniformly-sampled frames of
one generated video in temporal order. Judge the physics of the video
itself. Be conservative: reserve 5 for clearly flawless physics and 1 for
clearly broken physics.
\end{Verbatim}

\paragraph{User prompt.}
\begin{Verbatim}[breaklines,breaksymbolleft={}]
Task: Judge the PHYSICAL REALISM of this AI-generated robot /
embodied-manipulation video, from the video alone (ignore any task
instruction).

Criteria (your reasoning must address each; you may also note other
issues):
1. Agent integrity - the arm/gripper/hand stays structurally complete and
   consistent (no melting, fused/extra fingers, warping).
2. Scene & object consistency - background and objects stay temporally
   stable (no flicker, teleport, morphing, appear/disappear).
3. Interaction realism - contacts obey physics (grasps close and bear
   weight, no interpenetration, motion respects gravity/inertia).

Score (integer 1-5): 1 = gross violations throughout;
2 = major violations; 3 = noticeable local inconsistencies;
4 = minor issues only; 5 = no visible violation.

Reason first, then score. Output JSON only:
{"reasoning": "<assess agent integrity, scene & object consistency, and
 interaction realism, each with concrete visual evidence>",
 "physical_adherence": <1-5>}
\end{Verbatim}

\subsubsection{Instruction Alignment}

\paragraph{System prompt.}
\begin{Verbatim}[breaklines,breaksymbolleft={}]
You are a strict, calibrated evaluator of whether an AI-generated
embodied-manipulation video correctly performs a given task instruction.
You are shown the instruction and uniformly-sampled frames of one
generated video in temporal order; the FIRST frame is the initial scene
the video was conditioned on. Judge task execution, not raw visual
quality. Be conservative: reserve 5 for full, correct task completion and
1 for unrelated videos.
\end{Verbatim}

\paragraph{User prompt.}
\begin{Verbatim}[breaklines,breaksymbolleft={}]
Task: Judge whether this AI-generated video performs the instructed
manipulation task. The first frame is the initial scene the video was
conditioned on.

Instruction: "{instruction}"

Criteria (your reasoning must address each; you may also note other
issues):
1. Agent match - the task is done by the SAME manipulator shown in the
   first frame (not a different/new agent).
2. Object correctness - the manipulated object is the instruction's target
   object.
3. Goal completion - the instructed goal is actually achieved by the end
   (not merely approached).

Score (integer 1-5): 1 = unrelated or task not performed;
2 = major misalignment; 3 = partial completion;
4 = minor shortfalls only; 5 = full, correct completion.

Reason first, then score. Output JSON only:
{{"reasoning": "<assess agent match, object correctness, and goal
 completion, each with concrete evidence>",
 "instruction_alignment": <1-5>}}
\end{Verbatim}

%% file: appendix/non_mllm_metrics.tex
\subsection{Non-VLM Metrics}
\label{app:nonvlm_metrics}

We provide implementation details for the non-MLLM metric baselines reported in
Sec.~\ref{sec:evaluated_models}. These metrics quantify
complementary aspects of image quality, temporal consistency, motion
characteristics, and video-representation similarity without performing
MLLM-based semantic judgment.

All metrics are evaluated on the \textsc{RoboJudge-Bench} test set and aligned
with human annotations through the canonical \texttt{item\_id}. Unless
otherwise specified, each metric produces one scalar value for each generated
video. For raw distance- or error-based metrics, we reverse the score direction
when computing correlations such that larger values consistently correspond
to higher quality or stronger consistency.

The model-based baselines use the checkpoints configured in our evaluation
pipeline: CLIP ViT-B/32~\citep{radford2021clip},
DINO ViT-B/16~\citep{caron2021dino},
MUSIQ-SPAQ~\citep{ke2021musiq},
RAFT-Things~\citep{teed2020raft}, and a
V-JEPA~2 ViT-G checkpoint~\citep{assran2025vjepa2}
(\texttt{vjepa2-vitg-fpc64-384-ssv2}).

\paragraph{Uniform Random.}
We include a uniform random predictor as a trivial reference.
For each benchmark item, a score is independently sampled from the five-point
rating space,
\begin{equation}
    \hat{s}
    \sim
    \mathrm{Uniform}\{1,2,3,4,5\},
\end{equation}
for both Physical Adherence (PA) and Instruction Alignment (IA).
The resulting predictions are evaluated using the same agreement metrics as
the learned judges. For a fixed five-class human label, the expected
exact-match probability of the uniform predictor is $1/5=0.2$.

\paragraph{Image Quality with MUSIQ.}
We use MUSIQ~\citep{ke2021musiq} as a no-reference image-quality estimator,
following its use for imaging-quality evaluation in
VBench~\citep{huang2024vbench}. Our implementation uses the MUSIQ-SPAQ
checkpoint.

For a video
\begin{equation}
    V=\{I_t\}_{t=1}^{T},
\end{equation}
MUSIQ predicts a technical-quality score $q(I_t)$ independently for each
sampled frame. The video-level image-quality score is the temporal average
\begin{equation}
    S_{\mathrm{MUSIQ}}(V)
    =
    \frac{1}{T}
    \sum_{t=1}^{T}
    q(I_t).
\end{equation}

Following the VBench preprocessing protocol, frames are resized while
preserving aspect ratio when necessary, with the longer side capped at
512 pixels, and pixel intensities are normalized before inference.
A larger value indicates higher predicted frame-level technical quality.
MUSIQ evaluates individual-frame appearance and does not explicitly model
temporal consistency, physical plausibility, or task completion.

\paragraph{Pixel Frame Consistency.}
We construct a simple pixel-space temporal baseline. Videos are decoded at
8 fps, resized to $256\times256$, and converted to grayscale. Let
\begin{equation}
    G_t \in [0,255]^{H\times W}
\end{equation}
denote the processed frame at time $t$. For every pair of adjacent frames,
we compute the mean absolute pixel difference
\begin{equation}
    d_t^{\mathrm{pix}}
    =
    \frac{1}{HW}
    \left\|
        G_{t+1}-G_t
    \right\|_1.
\end{equation}

The video-level frame difference is
\begin{equation}
    D_{\mathrm{pix}}(V)
    =
    \frac{1}{T-1}
    \sum_{t=1}^{T-1}
    d_t^{\mathrm{pix}}.
\end{equation}

A lower value indicates stronger local frame consistency. For correlation
analysis, we reverse the metric direction so that larger values correspond to
stronger consistency. Affine rescaling does not change either Pearson or
Spearman correlation.

This metric is closely related to frame-difference-based temporal evaluation
used in video benchmarks such as VBench~\citep{huang2024vbench}. It captures
low-level visual stability and may penalize legitimate camera or object motion
even when the underlying dynamics are physically valid.

\paragraph{CLIP Frame Consistency.}
We use CLIP~\citep{radford2021clip} ViT-B/32 features to measure global
appearance consistency across a generated video, following the
feature-consistency formulation used for background consistency in
VBench~\citep{huang2024vbench}.

For each frame $I_t$, the CLIP image encoder produces an L2-normalized feature
vector
\begin{equation}
    \mathbf{c}_t
    =
    \frac{
        f_{\mathrm{CLIP}}(I_t)
    }{
        \|f_{\mathrm{CLIP}}(I_t)\|_2
    }.
\end{equation}

For every frame $t>1$, we compute its similarity to both the immediately
preceding frame and the first frame:
\begin{align}
    s_t^{\mathrm{prev}}
    &=
    \max
    \left(
        0,
        \mathbf{c}_{t-1}^{\top}\mathbf{c}_t
    \right),
    \\
    s_t^{\mathrm{first}}
    &=
    \max
    \left(
        0,
        \mathbf{c}_{1}^{\top}\mathbf{c}_t
    \right).
\end{align}

The frame-level consistency score is
\begin{equation}
    s_t^{\mathrm{CLIP}}
    =
    \frac{
        s_t^{\mathrm{prev}}
        +
        s_t^{\mathrm{first}}
    }{2},
\end{equation}
and the video-level score is
\begin{equation}
    S_{\mathrm{CLIP}}(V)
    =
    \frac{1}{T-1}
    \sum_{t=2}^{T}
    s_t^{\mathrm{CLIP}}.
\end{equation}

A higher score indicates stronger global visual consistency over time.
Although this protocol is commonly referred to as background consistency in
VBench, the feature extractor operates on the complete frame rather than an
explicitly segmented background region.

\paragraph{DINO Frame Consistency.}
We analogously use DINO~\citep{caron2021dino} ViT-B/16 features, following
the subject-consistency formulation of VBench~\citep{huang2024vbench}.
Let
\begin{equation}
    \mathbf{d}_t
    =
    \frac{
        f_{\mathrm{DINO}}(I_t)
    }{
        \|f_{\mathrm{DINO}}(I_t)\|_2
    }
\end{equation}
denote the normalized DINO representation of frame $I_t$.

For every $t>1$, the frame-level score is
\begin{equation}
    s_t^{\mathrm{DINO}}
    =
    \frac{1}{2}
    \left[
        \max
        \left(
            0,
            \mathbf{d}_{t-1}^{\top}\mathbf{d}_t
        \right)
        +
        \max
        \left(
            0,
            \mathbf{d}_{1}^{\top}\mathbf{d}_t
        \right)
    \right],
\end{equation}
and the video-level score is
\begin{equation}
    S_{\mathrm{DINO}}(V)
    =
    \frac{1}{T-1}
    \sum_{t=2}^{T}
    s_t^{\mathrm{DINO}}.
\end{equation}

Comparing each frame with both its immediate predecessor and the initial frame
captures short-term continuity while also penalizing longer-term identity or
appearance drift.

\paragraph{Optical-Flow Motion Magnitude.}
To characterize the amount of motion in a generated video, we estimate dense
optical flow using RAFT~\citep{teed2020raft}, following the flow-based motion
evaluation adopted in EvalCrafter~\citep{liu2024evalcrafter}. We use the
RAFT-Things checkpoint.

For consecutive frames $(I_t,I_{t+1})$, RAFT predicts a dense two-dimensional
flow field
\begin{equation}
    \mathbf{F}_t(x,y)
    =
    \left(
        u_t(x,y),
        v_t(x,y)
    \right).
\end{equation}

The mean motion magnitude for a frame pair is
\begin{equation}
    m_t
    =
    \frac{1}{HW}
    \sum_{x,y}
    \sqrt{
        u_t(x,y)^2
        +
        v_t(x,y)^2
    },
\end{equation}
and the video-level motion magnitude is
\begin{equation}
    S_{\mathrm{motion}}(V)
    =
    \frac{1}{T-1}
    \sum_{t=1}^{T-1}
    m_t.
\end{equation}

Motion magnitude has no intrinsic higher-is-better interpretation. A large
value indicates more apparent motion, but the motion may be either physically
plausible or erroneous. We therefore retain the original metric direction when
computing correlations with human ratings.

\paragraph{Optical-Flow Warp Error.}
We additionally compute a motion-compensated temporal consistency metric using
dense Farneb\"ack optical flow~\citep{farneback2003}. The preprocessing
follows the Pixel baseline: videos are decoded at 8 fps, resized to
$256\times256$, and converted to grayscale.

For two adjacent grayscale frames, we estimate
\begin{equation}
    \mathbf{F}^{\mathrm{FB}}_t
    =
    \mathrm{Farneback}
    \left(
        G_t,G_{t+1}
    \right),
\end{equation}
and warp $G_t$ toward $G_{t+1}$ using the estimated flow:
\begin{equation}
    \widetilde{G}_{t+1}
    =
    \mathcal{W}
    \left(
        G_t,
        \mathbf{F}^{\mathrm{FB}}_t
    \right),
\end{equation}
where $\mathcal{W}(\cdot)$ denotes dense image warping.

The residual photometric error is
\begin{equation}
    e_t
    =
    \frac{1}{HW}
    \left\|
        \widetilde{G}_{t+1}
        -
        G_{t+1}
    \right\|_1,
\end{equation}
and the video-level warp error is
\begin{equation}
    E_{\mathrm{warp}}(V)
    =
    \frac{1}{T-1}
    \sum_{t=1}^{T-1}
    e_t.
\end{equation}

A lower residual indicates that the observed transition is better explained
by a coherent motion field. For correlation analysis, we reverse the metric
direction such that larger values correspond to stronger motion-compensated
temporal consistency.

\paragraph{V-JEPA Paired Similarity.}
We construct a representation-level similarity metric using the V-JEPA~2
video encoder configured in our evaluation pipeline. For each generated
rollout $V_i^{\mathrm{gen}}$, we identify the corresponding real reference
trajectory $V_i^{\mathrm{gt}}$ from the same source dataset, task, and episode.

The generated and reference videos are processed using the same temporal
sampling and feature-extraction pipeline to obtain video-level embeddings
\begin{equation}
    \mathbf{z}^{\mathrm{gen}}_i
    =
    f_{\mathrm{VJEPA2}}
    \left(
        V_i^{\mathrm{gen}}
    \right),
    \qquad
    \mathbf{z}^{\mathrm{gt}}_i
    =
    f_{\mathrm{VJEPA2}}
    \left(
        V_i^{\mathrm{gt}}
    \right).
\end{equation}

The primary paired metric is cosine similarity:
\begin{equation}
    S_{\mathrm{pair}}(i)
    =
    \frac{
        \left(
            \mathbf{z}^{\mathrm{gen}}_i
        \right)^{\top}
        \mathbf{z}^{\mathrm{gt}}_i
    }{
        \left\|
            \mathbf{z}^{\mathrm{gen}}_i
        \right\|_2
        \left\|
            \mathbf{z}^{\mathrm{gt}}_i
        \right\|_2
    }.
\end{equation}

A larger value indicates that the generated rollout is closer to its paired
real trajectory in the learned spatiotemporal representation space.

We additionally compute feature mean-squared error as an auxiliary diagnostic:
\begin{equation}
    E_{\mathrm{pair}}(i)
    =
    \frac{1}{D}
    \left\|
        \mathbf{z}^{\mathrm{gen}}_i
        -
        \mathbf{z}^{\mathrm{gt}}_i
    \right\|_2^2,
\end{equation}
where $D$ is the embedding dimension. Since cosine similarity and feature MSE
provide closely related representation-level signals, the main comparison
reports the cosine-based Paired Similarity.

\paragraph{V-JEPA Manifold Similarity.}
Paired Similarity requires a corresponding real reference trajectory. We
therefore additionally construct an unpaired representation-level metric that
measures the proximity of each generated video to the feature distribution of
real embodied videos.

For manifold evaluation, each video is uniformly sampled to 16 frames and
resized to $128\times128$ before feature extraction. If multiple benchmark
items correspond to the same real source trajectory, that ground-truth video
contributes only once to the reference feature set. :contentReference[oaicite:2]{index=2}

Let
\begin{equation}
    \mathcal{Z}_{\mathrm{real}}
    =
    \left\{
        \mathbf{z}^{\mathrm{gt}}_1,
        \ldots,
        \mathbf{z}^{\mathrm{gt}}_M
    \right\}
\end{equation}
denote the set of unique real-video embeddings. We estimate the empirical mean
\begin{equation}
    \boldsymbol{\mu}
    =
    \frac{1}{M}
    \sum_{j=1}^{M}
    \mathbf{z}^{\mathrm{gt}}_j
\end{equation}
and covariance
\begin{equation}
    \boldsymbol{\Sigma}
    =
    \frac{1}{M-1}
    \sum_{j=1}^{M}
    \left(
        \mathbf{z}^{\mathrm{gt}}_j-\boldsymbol{\mu}
    \right)
    \left(
        \mathbf{z}^{\mathrm{gt}}_j-\boldsymbol{\mu}
    \right)^{\top}.
\end{equation}

To improve numerical stability, we regularize the covariance as
\begin{equation}
    \widetilde{\boldsymbol{\Sigma}}
    =
    \boldsymbol{\Sigma}
    +
    10^{-3}\mathbf{I},
\end{equation}
and use its Moore--Penrose pseudo-inverse. For a generated-video representation
$\mathbf{z}^{\mathrm{gen}}_i$, the raw manifold distance is the regularized
Mahalanobis distance
\begin{equation}
    d_i
    =
    \sqrt{
        \left(
            \mathbf{z}^{\mathrm{gen}}_i-\boldsymbol{\mu}
        \right)^{\top}
        \widetilde{\boldsymbol{\Sigma}}^{\dagger}
        \left(
            \mathbf{z}^{\mathrm{gen}}_i-\boldsymbol{\mu}
        \right)
    }.
\end{equation}

A smaller Mahalanobis distance indicates greater proximity to the real-video
feature manifold. For correlation analysis, we reverse the metric direction
so that larger values correspond to stronger manifold similarity.

Unlike distribution-level video metrics, this construction produces one
score for every generated rollout and can therefore be compared directly with
the corresponding human PA and IA ratings at the item level.

%% file: appendix/data_statistics.tex
\section{Data Statistics}
\label{appendix:data_statistics}
\input{tables/video_source_statistics}

%% file: tables/video_source_statistics.tex
\begin{table}[h]
\centering
\caption{\textbf{Video sources of \textsc{RoboJudge-Train} and \textsc{RoboJudge-Bench}.}
The training set contains 11,520 generated videos from 14 generators and 831 verified successful real demonstrations paired with their corresponding instructions; the latter are assigned PA=5 and IA=5. The test set contains 800 videos from 10 of these generators; ``--'' indicates a generator not included in the test set.}
\label{tab:appendix_train_data}
\small
\setlength{\tabcolsep}{4.5pt}
\begin{tabular}{l r r}
\toprule
\textbf{Video Generation Model / Source} & \textbf{Train} & \textbf{Test} \\
\midrule
CogVideo I2V              & 2,099 & 96  \\
GigaWorld          & 1,550 & 105 \\
WoW Wan14B              & 1,533 & 122 \\
ABOT PhysWorld            & 1,473 & 94  \\
Cosmos3-Nano                        & 1,033 & 107 \\
Wan2.2 TI2V-5B
                            & 971 & 110 \\
Wan2.6 Flash                   & 781 & 69  \\
PF Wan14B                  & 551 & --  \\
Vidu Q3                        & 280 & --  \\
LingBot Video MoE Base         & 277 & --  \\
Veo 3.1 Lite                   & 274 & 35  \\
HunyuanVideo 1.5               & 248 & --  \\
HappyHorse                     & 231 & 36  \\
Kling2.2                          & 219 & 26  \\
\midrule
\textbf{Generated videos}      & \textbf{11,520} & \textbf{800} \\
GT real videos    & 831 & -- \\
\midrule
\textbf{Total videos}          & \textbf{12,351} & \textbf{800} \\
\midrule
\multicolumn{3}{l}{\textit{Axis-level training samples: 12,351 PA + 12,351 IA = 24,702}} \\
\bottomrule
\end{tabular}
\end{table}

%% file: appendix/video_generation_prompt_engineering.tex
\section{Video Prompt Engineering}
\label{appendix:video_prompt_engineering}

We provide the complete prompt-engineering procedure used to transform source manipulation instructions into prompts for video generation. Given an initial frame and an instruction, the actual model input is constructed as

\begin{quote}
\texttt{``The attached image is the first frame of a \{kind\} manipulation video.''}\\
\texttt{+ rewrite template}\\
\texttt{+ ``If the instruction conflicts with the image, prioritize the image.''}\\
\texttt{+ ``Output ONLY the rewritten prompt.''}\\
\texttt{+ ``Instruction: '' + original instruction,}
\end{quote}

where \texttt{\{kind\}} is either \texttt{robot} or \texttt{human}. The initial frame is provided as the image input to the rewriting model.

\subsection{Instruction Rewriting}

We use five rewriting templates for each manipulation type. All templates impose a 100-word upper bound (approximately 128 tokens) and encourage the rewritten prompt to preserve the object name from the original instruction.

\subsubsection{Robot Rewriting Templates}

\begin{enumerate}
\item \texttt{Rewrite the instruction as a concise text-to-video prompt (<=100 words, \string~128 token hard cap) describing the robot/gripper performing the task action on the locked object (use the object name from the instruction verbatim), through to the completion state. Plain narrative. Keep scene description minimal - focus on what the robot is doing.}

\item \texttt{Rewrite the instruction as a text-to-video prompt (<=100 words, \string~128 token hard cap) breaking the task into reach -> grasp -> execute -> release/complete on the locked object (object name verbatim), ending in the completion state. Mention scene only briefly if needed for grounding.}

\item \texttt{Rewrite the instruction as an imperative text-to-video prompt (<=100 words, \string~128 token hard cap) commanding the robot/gripper to execute the task on the locked object (object name verbatim) and finish in the completion state. Use action verbs. Stay action-centric; keep environmental detail spare.}

\item \texttt{Rewrite the instruction as a text-to-video prompt (<=100 words, \string~128 token hard cap) starting with the action verb, unfolding what the robot/gripper does to the locked object (object name verbatim) through to completion. Keep background mentions brief.}

\item \texttt{Rewrite the instruction as a text-to-video prompt (<=100 words, \string~128 token hard cap) decomposing into approach -> contact -> completion sub-actions performed by the robot/gripper on the locked object (object name verbatim), ending at the final state. Restrain scene/background detail.}

\end{enumerate}

\subsubsection{Human Rewriting Templates}

\begin{enumerate}
\item \texttt{Rewrite the instruction as a concise text-to-video prompt (<=100 words, \string~128 token hard cap) describing the person's hand performing the task action on the locked object (use the object name from the instruction verbatim), through to the completion state. Plain narrative. Keep scene description minimal - focus on what the hand is doing.}

\item \texttt{Rewrite the instruction as a text-to-video prompt (<=100 words, \string~128 token hard cap) breaking the task into reach -> grasp -> execute -> release/complete on the locked object (object name verbatim), ending in the completion state. Mention scene only briefly if needed for grounding.}

\item \texttt{Rewrite the instruction as an imperative text-to-video prompt (<=100 words, \string~128 token hard cap) commanding the person's hand to execute the task on the locked object (object name verbatim) and finish in the completion state. Use action verbs. Stay action-centric; keep environmental detail spare.}

\item \texttt{Rewrite the instruction as a text-to-video prompt (<=100 words, \string~128 token hard cap) starting with the action verb, unfolding what the person's hand does to the locked object (object name verbatim) through to completion. Keep background mentions brief.}

\item \texttt{Rewrite the instruction as a text-to-video prompt (<=100 words, \string~128 token hard cap) decomposing into approach -> contact -> completion sub-actions performed by the person's hand on the locked object (object name verbatim), ending at the final state. Restrain scene/background detail.}

\end{enumerate}

\subsection{Prompt Prefixing}

As an alternative to instruction rewriting, we prepend a fixed contextual prefix to the original instruction. The prefix is selected according to the manipulation type and is not passed through the rewriting model.

\subsubsection{Robot Prefixes}

\begin{enumerate}
\item \texttt{In a fixed robotic workspace, a rigid and physically consistent embodied robotic arm with high stability and no deformation proceeds to}
\item \texttt{In a controlled robotic workspace, a stable rigid robotic arm with no deformation begins to}
\item \texttt{Within a fixed embodied manipulation setup, a physically consistent robotic arm proceeds to}
\item \texttt{In a robotic workspace, a stable robotic arm}
\item \texttt{A rigid, physically plausible robotic arm in the fixed workspace proceeds to}
\end{enumerate}

\subsubsection{Human Prefixes}

\begin{enumerate}
\item \texttt{In a first-person egocentric video, a real human hand starts to}
\item \texttt{From an egocentric point of view, the person's hand naturally proceeds to}
\item \texttt{In a real-world first-person scene, a human hand begins to}
\item \texttt{From a head-mounted camera, the person's bare hand starts to}
\item \texttt{In an egocentric recording of a person, their hand proceeds to}
\end{enumerate}

For prefix-based prompts, the original instruction is appended directly after the selected prefix without an additional rewriting step. Thus, prompt variants preserve the original task semantics while introducing controlled variation in the description of the acting entity and visual context.

%% file: appendix/additional_analysis.tex
\section{Additional Analysis}
\label{appendix:additional_analysis}

\begin{figure}[t]
\centering
\includegraphics[width=1.0\linewidth]{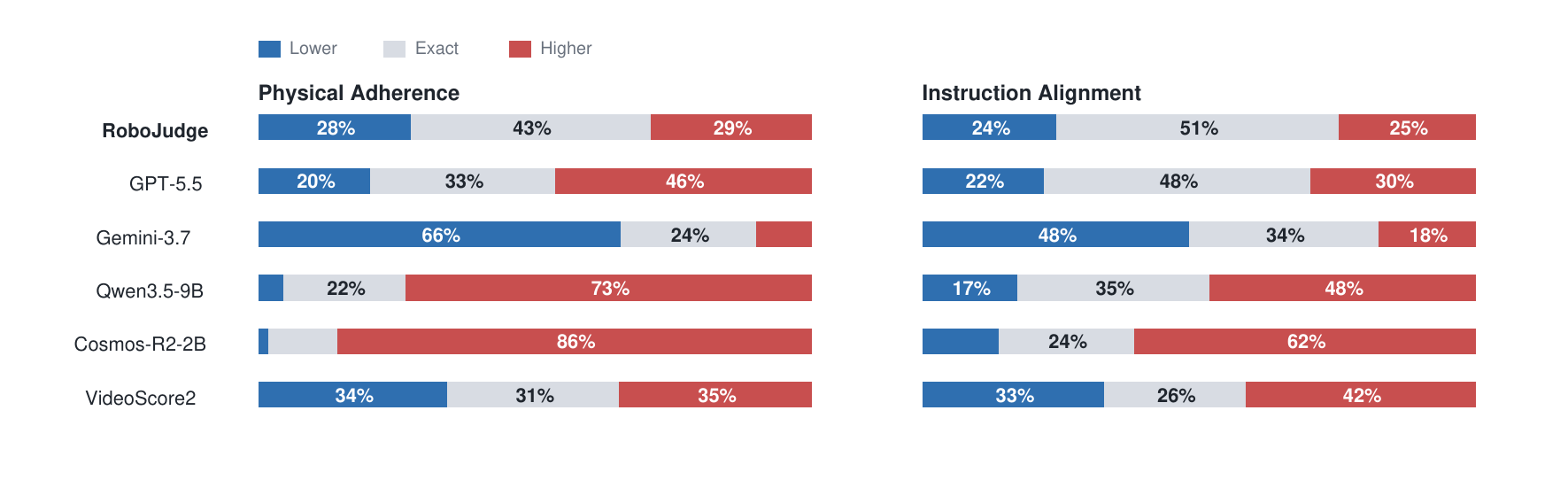}
\caption{Directional scoring bias relative to human ratings. RoboJudge is balanced around human scores, while baselines tend to over-score or under-score.}
\label{fig:signed_bias}
\end{figure}

Figure~\ref{fig:signed_bias} shows that several baselines have clear directional bias: on PA, Qwen3.5-9B and Cosmos-R2-2B over-score 73.3\% and 85.8\% of examples, while Gemini-3.7 under-scores 65.5\%. RoboJudge is more balanced on both PA (29.1\% higher vs. 27.5\% lower) and IA (24.8\% vs. 24.2\%).

\subsection{Statistical Confidence}
\label{appendix:bootstrap}

We assess the robustness of RoboJudge-9B's gains using a paired cluster
bootstrap with 10,000 replicates (random seed 20260926). We treat the 558 source
clips, identified by source corpus, task, and episode, as clusters and jointly
resample all generated videos derived from each clip. Sampling is stratified by
the eight source corpora, and each replicate uses the same sampled clips for
RoboJudge-9B and the comparison model. We report percentile 95\% confidence intervals for the
difference in Pearson correlation. Overall correlation follows the definition in
Table~\ref{tab:judge_performance}, pooling PA and IA axis-level predictions.

\begin{table}[t]
\centering
\caption{Paired cluster-bootstrap comparisons between \textsc{RoboJudge-9B}
and the strongest model baseline for each metric. $\Delta r$ denotes
RoboJudge-9B minus the comparison model.}
\label{tab:bootstrap_significance}
\small
\setlength{\tabcolsep}{5pt}
\begin{tabular}{l l cc}
\toprule
\textbf{Metric} & \textbf{Strongest baseline} & $\boldsymbol{\Delta r}$ & \textbf{95\% CI} \\
\midrule
PA & Gemini-3.7-Flash & $+0.034$ & $[-0.011,\ 0.078]$ \\
IA & GPT-5.5 & $+0.044$ & $[0.010,\ 0.078]$ \\
Overall & Gemini-3.7-Flash & $+0.089$ & $[0.061,\ 0.119]$ \\
\bottomrule
\end{tabular}
\end{table}

RoboJudge-9B improves over the strongest baseline in all three point estimates.
The bootstrap intervals confirm robust gains in IA and
overall agreement, while RoboJudge-9B also achieves the highest observed PA
correlation.

\subsection{Video Generation Leaderboard}
\label{appendix:videogen_leaderboard}

Table~\ref{tab:videogen_leaderboard} reports the full \textsc{RoboJudge-9B} scores of the video generation models ranked in Figure~\ref{fig:videogen_leaderboard} and discussed in Sec.~\ref{sec:videogen_leaderboard}.

\input{tables/video_generation_leaderboard}

\subsection{Judge Agreement Landscape}
\label{appendix:judge_landscape}

\begin{figure}[H]
\centering
\includegraphics[width=0.5\linewidth]{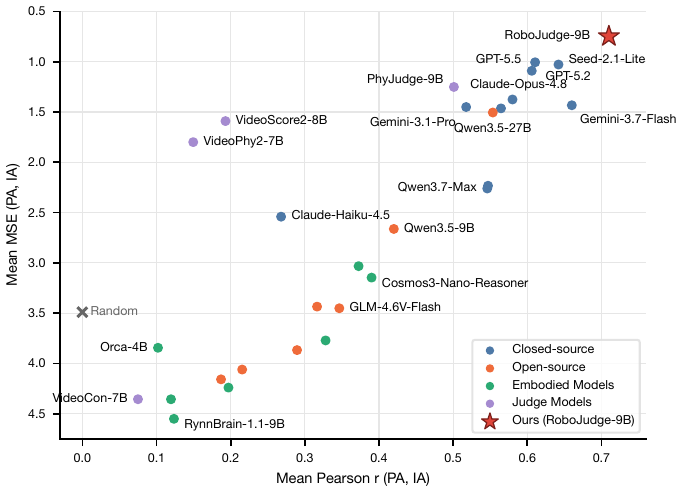}
\caption{Judge agreement with human ratings on \textsc{RoboJudge-Bench}.}
\label{fig:judge_landscape}
\end{figure}

Figure~\ref{fig:judge_landscape} plots each judge's mean Pearson correlation with human PA and IA ratings against its mean squared error, complementing Table~\ref{tab:judge_performance} in Sec.~\ref{sec:overall_judge}.

%% file: tables/video_generation_leaderboard.tex
\begin{table}[H]
\centering
\caption{\textbf{RoboJudge-9B evaluation results on generated videos.}
PA denotes Physical Adherence, IA denotes Instruction Alignment, and
Overall is the average of PA and IA.}
\label{tab:videogen_leaderboard}
\footnotesize
\setlength{\tabcolsep}{5pt}
\begin{tabular}{@{}l ccc@{}}
\hline
\textbf{Video Generation Model} & \textbf{PA} $\uparrow$ & \textbf{IA} $\uparrow$ & \textbf{Overall} $\uparrow$ \\
\midrule

HappyHorse
& 3.821 & 4.313 & 4.067 \\

Wan2.6 Flash
& 3.827 & 4.027 & 3.927 \\

PF Wan14B
& 3.680 & 3.860 & 3.770 \\

Vidu Q3
& 3.480 & 3.940 & 3.710 \\

Kling2.2
& 3.420 & 3.940 & 3.680 \\

Veo 3.1 Lite
& 3.580 & 3.780 & 3.680 \\

Cosmos3-Super
& 3.372 & 3.623 & 3.497 \\

ABOT PhysWorld
& 3.415 & 3.458 & 3.437 \\

Cosmos3-Nano
& 3.350 & 3.500 & 3.425 \\

LingBot Video MoE Base
& 3.960 & 2.640 & 3.300 \\

HunyuanVideo 1.5
& 3.220 & 3.060 & 3.140 \\

LongCat
& 3.440 & 2.620 & 3.030 \\

GigaWorld
& 2.781 & 2.994 & 2.887 \\

WoW Wan14B
& 2.860 & 2.733 & 2.797 \\

SVD-XT
& 3.360 & 1.940 & 2.650 \\

Wan2.2 TI2V-5B
& 2.656 & 2.512 & 2.584 \\

CogVideo I2V
& 2.014 & 2.151 & 2.082 \\

\hline
\end{tabular}
\end{table}

%% file: appendix/additional_examples.tex
\section{Additional Examples of \textsc{RoboJudge-Bench}}
\label{appendix:more_examples}

We present additional examples from \textsc{RoboJudge-Bench} across diverse
embodiments, source datasets, and video generation models. Each example shows
representative video frames, PA/IA scores, and the corresponding fine-grained
rationales.

\vspace{0.4em}

\noindent
\includegraphics[
    page=1,
    width=\linewidth,
    trim=24 28 24 28,
    clip
]{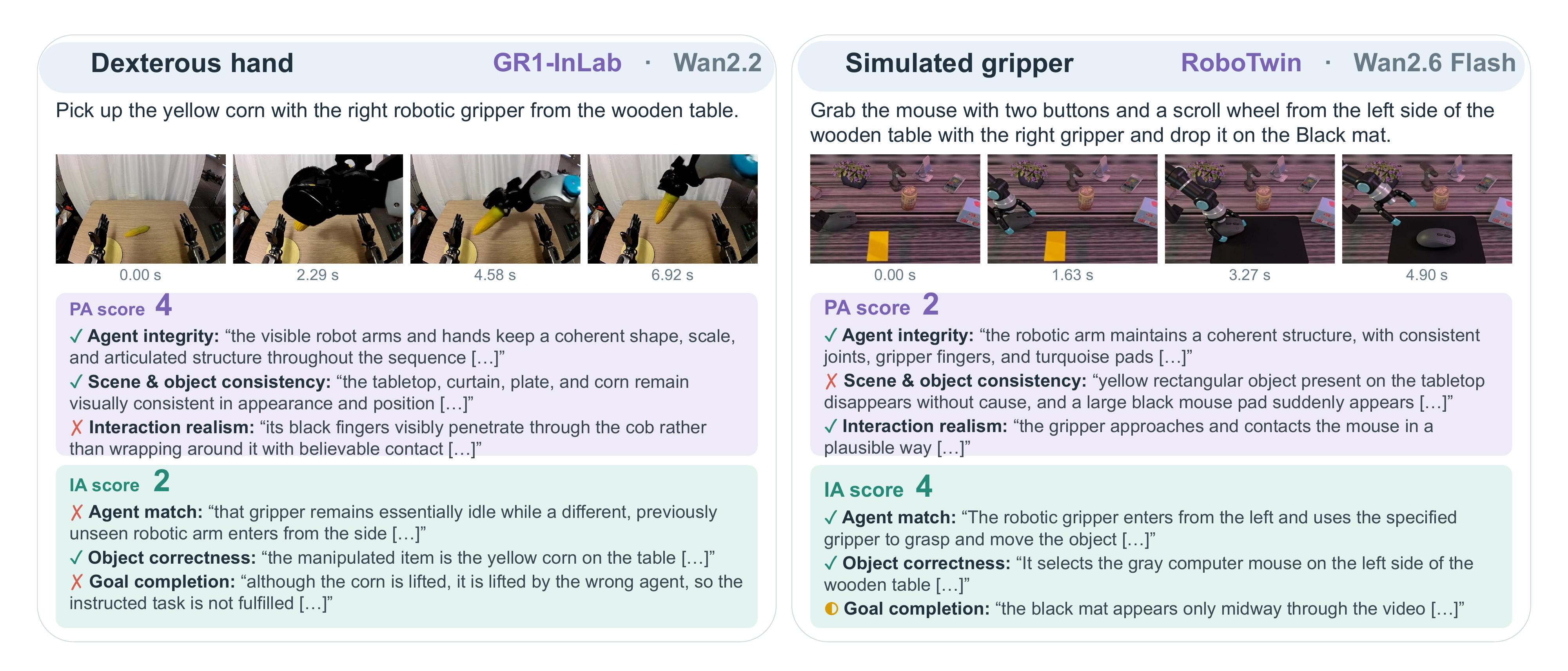}

\vspace{-0.25em}

\noindent
\includegraphics[
    page=2,
    width=\linewidth,
    trim=24 28 24 28,
    clip
]{figures/additional_examples.pdf}

\vspace{0.2em}

\refstepcounter{figure}
\label{fig:additional_robojudge_examples}
\noindent
\textbf{Figure \thefigure:}
Additional qualitative examples from \textsc{RoboJudge-Bench}, covering
diverse embodiments, source datasets, generation models, and PA/IA score
combinations. Each case includes fine-grained rationales for the corresponding
PA and IA judgments.

\vspace{0.8em}

\noindent
\includegraphics[
    page=3,
    width=\linewidth,
    trim=24 28 24 28,
    clip
]{figures/additional_examples.pdf}

\vspace{-0.25em}

\noindent
\includegraphics[
    page=4,
    width=\linewidth,
    trim=24 28 24 28,
    clip
]{figures/additional_examples.pdf}

\vspace{0.2em}

\noindent
\textbf{Figure \thefigure\ (continued):}
Additional qualitative examples from \textsc{RoboJudge-Bench}.

%% file: appendix/experiment_details.tex
\section{Experimental Details}
\label{appendix:exp_setup}

\subsection{Computation}

We use 4 NVIDIA H20 GPUs for training \textsc{RoboJudge-9B}. The SFT stage takes approximately 2 hours to complete one epoch. The subsequent GRPO stage is initialized from the SFT checkpoint and takes approximately 12 hours to reach 200 training steps.

For SFT, we perform full-parameter fine-tuning with a global batch size of 64 and a learning rate of $1\times10^{-5}$. For GRPO, we use a per-device batch size of 2 with gradient accumulation of 8 and 8 generations per prompt, corresponding to 64 prompts and 512 sampled completions per update. We use a learning rate of $1\times10^{-6}$ and a maximum completion length of 512 tokens.

\subsection{Training Log}

\begin{figure*}[t]
    \centering
    \begin{minipage}[t]{0.48\textwidth}
        \centering
        \includegraphics[width=\linewidth]{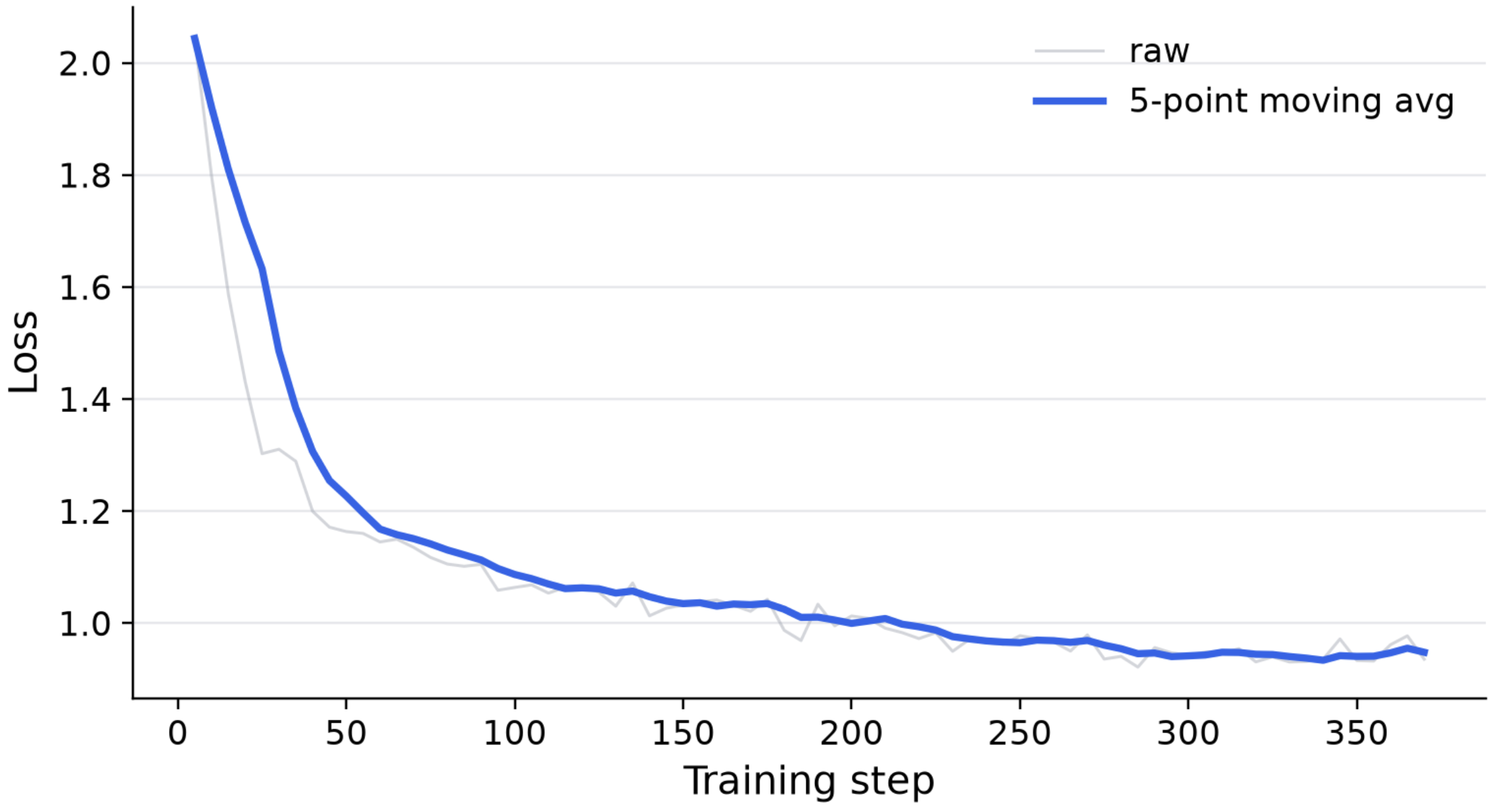}
        \caption{\textbf{SFT Training Loss.} Training loss throughout the SFT stage.}
        \label{fig:sft_loss}
    \end{minipage}
    \hfill
    \begin{minipage}[t]{0.48\textwidth}
        \centering
        \includegraphics[width=\linewidth]{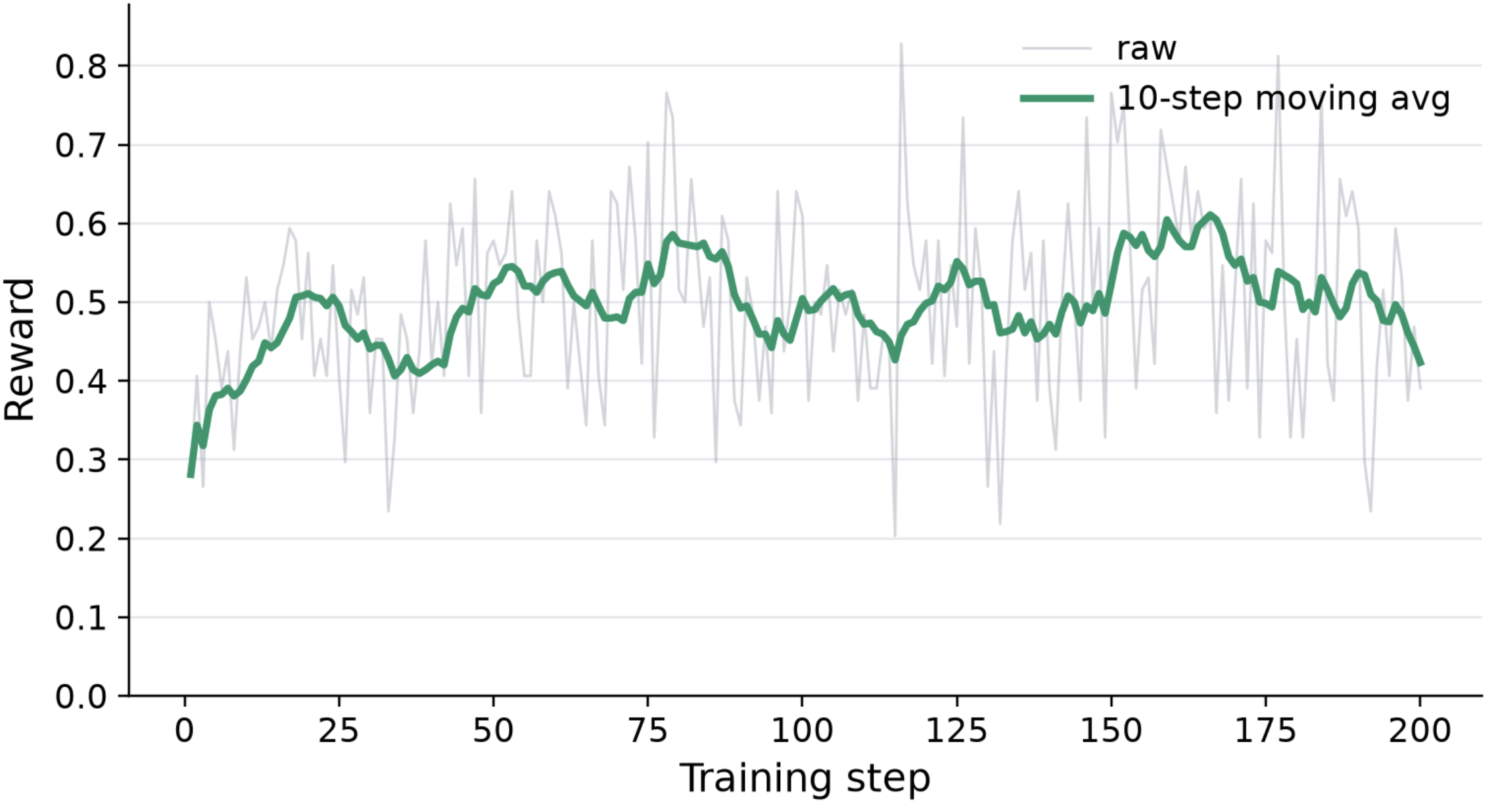}
        \caption{\textbf{GRPO Training Reward.} Training reward throughout the GRPO stage.}
        \label{fig:rl_reward}
    \end{minipage}
\end{figure*}
\subsection{Frame-Rate Ablation}
\label{appendix:fps_ablation}

Table~\ref{tab:fps_ablation} reports the full frame-rate ablation summarized in Sec.~\ref{sec:ablation}. The 4 FPS entries are the same canonical evaluations reported in the main results table.

\input{tables/frame_rate_ablation}

%% file: tables/frame_rate_ablation.tex
\begin{table}[H]
\centering
\caption{Frame-rate (FPS) ablation on \textsc{RoboJudge-Bench}. Acc is exact-match accuracy. The 4 FPS rows reuse the corresponding main-table results because 4 FPS is the default evaluation setting.}
\label{tab:fps_ablation}
\fontsize{7.5pt}{8.5pt}\selectfont
\setlength{\tabcolsep}{2.8pt}
\begin{tabular}{@{}l c ccc ccc ccc@{}}
\toprule
\multirow{2}{*}{\textbf{Judge}} &
\multirow{2}{*}{\textbf{FPS}} &
\multicolumn{3}{c}{\textbf{PA}} &
\multicolumn{3}{c}{\textbf{IA}} &
\multicolumn{3}{c}{\textbf{Overall}} \\
\cmidrule(lr){3-5}
\cmidrule(lr){6-8}
\cmidrule(lr){9-11}
& & $r\uparrow$ & Acc$\uparrow$ & MSE$\downarrow$ & $r\uparrow$ & Acc$\uparrow$ & MSE$\downarrow$ & $r\uparrow$ & Acc$\uparrow$ & MSE$\downarrow$ \\
\midrule

\multirow{3}{*}{GPT-5.5}
& 2 & 0.483 & 0.344 & 1.339 & 0.735 & 0.484 & 0.792 & 0.620 & 0.414 & 1.066 \\
& 4 & 0.489 & 0.334 & 1.359 & 0.733 & 0.481 & 0.786 & 0.618 & 0.407 & 1.073 \\
& 8 & 0.483 & 0.351 & 1.335 & 0.738 & 0.500 & 0.775 & 0.620 & 0.426 & 1.055 \\

\midrule

\multirow{3}{*}{Qwen3.5-9B}
& 2 & 0.316 & 0.214 & 3.422 & 0.429 & 0.355 & 2.522 & 0.363 & 0.285 & 2.972 \\
& 4 & 0.379 & 0.223 & 3.011 & 0.461 & 0.347 & 2.314 & 0.401 & 0.285 & 2.663 \\
& 8 & 0.379 & 0.214 & 3.058 & 0.461 & 0.346 & 2.320 & 0.401 & 0.280 & 2.689 \\

\bottomrule
\end{tabular}
\end{table}